\documentclass[11 pt, onecolumn]{article}

\usepackage[margin=1.1in]{geometry}

\usepackage{amsmath}
\usepackage{bm}	
\usepackage{graphicx}

\usepackage{graphicx}
\usepackage{multirow}
\usepackage{mdwlist}

\usepackage{threeparttable}

\usepackage{amsfonts}
\usepackage{amsmath}
\usepackage{hyperref}
\usepackage{bbm}
\usepackage{float}
\usepackage{caption}
\usepackage{subcaption}
\usepackage{xcolor}
\usepackage{mathtools}
\usepackage{tikz}

\usetikzlibrary{shapes.geometric,arrows}
\usepackage{enumerate}
\usepackage{multicol}
\usepackage{stackrel}

\usepackage{subcaption}
\usepackage{amssymb}
\usepackage[utf8]{inputenc}
\usepackage{booktabs,lipsum}
\usepackage{hyperref}
\usepackage{algorithm}
\usepackage{algpseudocode}
\usepackage{caption}

\DeclareMathAlphabet{\mathcal}{OMS}{cmsy}{m}{n}

\usepackage{graphicx}
\usepackage{epstopdf}

\begin{document}
\title{\LARGE \bf AI for Regulatory Affairs: Balancing Accuracy, Interpretability, and Computational Cost in Medical Device Classification}

\author{Yu Han, Aaron Ceross, and Jeroen H.M. Bergmann
\thanks{Yu Han is with the Institute of Biomedical Engineering, Department of Engineering Science, University of Oxford, Oxford, UK. Aaron Ceross is with Birmingham Law School, University of Birmingham, Birmingham, UK. Jeroen H.M. Bergmann is with the Institute of Biomedical Engineering, University of Oxford, and the Department of Technology and Innovation, University of Southern Denmark, Odense, Denmark, emails: yu.han@ox.ac.uk, a.ceross@bham.ac.uk, jeroen.bergmann@eng.ox.ac.uk}%
}
\newcommand*{\QEDA}{\hfill\ensuremath{\blacksquare}}%

\maketitle

\begin{abstract}

Artificial intelligence (AI) offers promising solutions for innovation and commercialization in healthcare, particularly in enhancing the efficiency and reliability of regulatory processes. Regulatory affairs, which sits at the intersection of medicine and law, can benefit significantly from AI-enabled automation. Classification task is the initial step in which manufacturers position their products to regulatory authorities, and it plays a critical role in determining market access, regulatory scrutiny, and ultimately, patient safety. In this study, we investigate a broad range of AI models—including traditional machine learning (ML) algorithms, deep learning architectures, and large language models (LLMs)—using a regulatory dataset of medical device descriptions. We evaluate each model along three key dimensions: classification accuracy, interpretability, and computational cost. Our analysis highlights the trade-offs between these metrics and emphasizes that the optimal model choice depends on the specific priorities of the end user—whether the need is for high accuracy, transparent decision-making, or computational efficiency. Our results show that traditional models such as XGBoost and Random Forest deliver strong accuracy and F1 scores while maintaining low inference latency, making them well-suited for deployment in practical settings. Logistic regression, while slightly less accurate, offers clear interpretability and exceptional computational speed. Deep learning models, particularly CNNs, outperform traditional approaches in accuracy but require modestly greater computational resources. In contrast, LLMs demonstrate lower classification performance and significantly higher inference times, though they provide the advantage of natural language-based explanations. No single model outperforms others across all evaluation criteria, highlighting the need for context-dependent, hybrid strategies. To support such decision-making, we present one of the first comprehensive benchmarks of AI models for medical device classification in a regulatory context, offering empirical insights and practical guidance for integrating AI into regulatory workflows.
\end{abstract}


\section{Introduction}
In recent years, regulatory agencies have shown growing interest in leveraging artificial intelligence (AI) to enhance the efficiency and accuracy of their review processes. Notably, the U.S. Food and Drug Administration (FDA) completed its first AI-assisted scientific review pilot in early 2025 and announced plans for agency-wide deployment of generative AI tools across all centers by mid 2025 ~\cite{FDA2025}. As part of this initiative, the FDA engaged in exploratory discussions with OpenAI on a project referred to as “CDERGPT,” aiming to incorporate large language models (LLMs) into product evaluation and regulatory workflows ~\cite{Heaven2025}. These developments underscore a broader global trend toward integrating AI into regulatory science, highlighting the urgent need for rigorous, context-sensitive evaluations of AI models—particularly in high-stakes domains such as medical device classification, where interpretability and legal compliance are paramount.

Regulatory affairs is a multidisciplinary field dedicated to ensuring that healthcare products—including medical devices, pharmaceuticals, and biologics—comply with the stringent requirements of government agencies and international health authorities. These agencies include the U.S. Food and Drug Administration (FDA), the European Medicines Agency (EMA), and the National Medical Products Administration (NMPA) in China, among others. Regulatory professionals are responsible for guiding products through a complex and evolving landscape of compliance, from early-stage development to clinical testing, market authorization, and post-market surveillance. Among the many responsibilities in this field, the classification of medical devices is foundational, as it determines the necessary level of regulatory scrutiny. Depending on the classification, the requirements for clinical evidence, labeling, safety documentation, and conformity assessments vary substantially~\cite{mantus2014fda, kumari2016current, han2024regulatory}.

Globally, the risk-based classification system remains the dominant framework for evaluating medical devices~\cite{han2023uncovering, muehlematter2023fda}. In the U.S., the FDA mandates that Class I devices undergo general controls, while Class II devices typically require 510(k) clearance, and Class III devices—representing the highest risk—must pass rigorous premarket approval (PMA). Similarly, under the EU's Medical Device Regulation (MDR), and comparable systems in Australia and China, increasing classification levels correspond to increased regulatory requirements. Non-compliance with these systems can result in legal and financial consequences, including registration failure, recalls, or product bans.

A crucial first step in this regulatory workflow is the submission of detailed product descriptions. These descriptions are central to the classification decision, as regulators typically rely on textual claims made by manufacturers to assess intended use, operational features, and risk levels~\cite{aronson2020medical, pomeranz2013comprehensive}. Misleading or incomplete disclosures can not only compromise patient safety but also breach legal requirements. For instance, if a device incorporates AI functionality that is not clearly stated in the product description, regulators will not classify it accordingly, and subsequent marketing claims regarding AI use would be considered non-compliant~\cite{herman2021position}. Thus, product classification hinges on accurate and complete textual information, which becomes a critical input for any AI-driven regulatory analysis. 


With the increasing complexity and volume of medical regulations, AI-driven approaches are becoming indispensable for regulatory affairs. Researchers have noted a steady annual increase in regulatory complexity, accompanied by an expansion in the number of required classifications and compliance assessments ~\cite{arnould2021complexity}, ~\cite{han2024more}. To address those, researchers have explored the application of machine learning (ML) and natural language processing (NLP) to automate text classification tasks. In clinical contexts, ML and deep learning models have proven effective for tasks such as diagnosis prediction, treatment recommendation, and report classification~\cite{yin2021role, castiglioni2021ai, imaichi2013comparison}. For regulatory affairs—where legal and scientific language intersects—the opportunity for automation is substantial but underexplored. Prior work has focused on specific use cases, such as AI-assisted dossier construction, data extraction from forms, and safety signal detection in post-market surveillance~\cite{patil2023artificial, khinvasara2024post}.

Text processing methods for medical product classification is still underexplored. Text processing task can be broadly categorized into rule-based and machine learning-based approaches. Rule-based methods rely on manually predefined classification rules ~\cite{kumar2010automatic}, which regulatory experts construct based on existing guidelines and domain knowledge. While such methods can achieve high accuracy in well-defined scenarios ~\cite{imaichi2013comparison}, they suffer from significant limitations, including high maintenance costs, lack of adaptability, and difficulty in handling complex, unstructured regulatory texts ~\cite{khan2010review}. Each update to regulatory guidelines necessitates manual revisions to rule sets, making them impractical for large-scale classification tasks. Additionally, reliance on human consultants introduces cost inefficiencies and potential inconsistencies, as manual classification decisions are susceptible to variations in expert judgment ~\cite{korger2021rule}. These challenges highlight the need for scalable, automated approaches capable of adapting to evolving regulatory requirements while maintaining interpretability and reliability.

Machine learning-based methods could automatically learning classification patterns from historical regulatory data. Traditional algorithms such as Support Vector Machines (SVMs) and Naïve Bayes classifiers have been widely used in text classification tasks. SVMs are particularly effective in high-dimensional spaces, leveraging hyperplane-based decision boundaries to separate different categories ~\cite{awad2015support}. Naïve Bayes classifiers, which assume feature independence, provide computational efficiency and robustness for large text datasets ~\cite{webb2010naive}, making them well-suited for initial classification tasks in regulatory affairs. Additionally, semantic analysis techniques have been employed to extract meaningful patterns from medical texts, improving classification accuracy and decision-making in regulatory applications ~\cite{salama2016semantic, zhang2016semantic}. However, traditional machine learning models typically rely on manual feature engineering, limiting their adaptability to diverse regulatory texts and reducing their ability to generalize across different classification scenarios. Recent work by Ceross et al. has shown that machine learning models can classify medical devices based on structured and unstructured data extracted from regulatory databases, such as the Australian Register of Therapeutic Goods (ARTG) ~\cite{ceross2021machine}.

Deep learning and transformer-based models have emerged as promising tools for complex NLP tasks, including classification, named entity recognition, and entailment analysis. Large language models (LLMs) are capable of understanding context-rich input, capturing domain-specific language, and generating human-like interpretations~\cite{banerjee2023large}. These models hold potential for more nuanced classification decisions, particularly when regulatory guidance is ambiguous or device descriptions are detailed and complex. Nevertheless, their adoption in regulatory environments remains limited due to concerns about interpretability, computational overhead, and legal accountability.

Artificial intelligence offers the potential to automate or assist in medical device classification, but regulatory applications demand more than just high accuracy. In this context, interpretability of AI models is crucial so that regulators can trust and understand the basis of a model’s recommendation ~\cite{wu2024regulating}, \cite{shen2023towards}. For example, an algorithm that classifies a device as high-risk should be able to explain whether it was due to invasive use, critical function, or other factors recognizable to regulatory scientists. Additionally, computational cost becomes important for practical adoption: regulatory agencies may need to run these models at scale or on secure systems with limited resources, making efficiency and speed necessary considerations. Balancing accuracy with interpretability and efficiency is therefore a key challenge for deploying AI in this domain~\cite{wang2025comprehensive}. The balance between these factors is not fixed, and regulatory decisions should be based on the particular context. 

In this study, we pick classification as RA downstream task, seek to systematically evaluate multiple AI approaches—ranging from rule-based to traditional ML, deep learning, and LLMs—for the task of medical device classification based on regulatory product descriptions. We examine three core challenges:

\begin{enumerate}
    \item \textbf{Accuracy:} How well can each model predict the correct regulatory class from free-text product descriptions?
    \item \textbf{Interpretability:} Can we explain why a model made a decision, and do these explanations align with regulatory reasoning?
    \item \textbf{Cost:} What is the computational cost of using high-performing models, particularly LLMs, and how does this impact scalability and sustainability?
\end{enumerate}

\section{Related Work}

Text classification has long been a central task in natural language processing (NLP), and a wide array of computational methods have been proposed to solve it, ranging from rule-based systems to machine learning (ML), deep learning (DL), and large language models (LLMs). Each category has shown strengths and limitations in regulatory applications such as medical device classification, particularly regarding interpretability, scalability, and computational efficiency.

\subsection{Rule-Based Systems}

Rule-based systems rely on manually crafted patterns, ontologies, or decision trees to categorize text data. These approaches have traditionally dominated regulatory domains due to their interpretability and adherence to predefined logic ~\cite{kumar2010automatic, korger2021rule}. For instance, keyword-based systems can classify medical device descriptions based on phrases like "non-invasive" or "implantable," which directly map to regulatory risk categories. However, they lack scalability and adaptability to novel or nuanced language use ~\cite{khan2010review, imaichi2013comparison}, requiring constant manual updates when regulatory guidelines evolve.

\subsection{Traditional Machine Learning}

Classic ML models such as Naive Bayes, Support Vector Machines (SVMs), Logistic Regression, and ensemble methods like Random Forest and XGBoost have been extensively used in text classification tasks ~\cite{webb2010naive, amari1999improving, chen2016xgboost}. These models rely on preprocessed and vectorized representations of text, such as TF-IDF or word embeddings, and provide strong performance when combined with effective feature engineering. Naive Bayes and Logistic Regression offer interpretable decision-making via conditional probabilities and coefficients, respectively, while SVMs and tree-based models require post-hoc explanation methods ~\cite{lundberg2017unified}.

XGBoost in particular has emerged as a highly effective model due to its ability to handle non-linear feature interactions, and is now a standard baseline in many structured and semi-structured classification tasks ~\cite{chen2016xgboost}. Recent studies in regulatory informatics have demonstrated the feasibility of using these models for classification based on structured product descriptions ~\cite{ceross2021machine}.

\subsection{Deep Learning Models}

Neural network-based models, especially Convolutional Neural Networks (CNNs) and Recurrent Neural Networks (RNNs), have been adopted for regulatory text classification for their ability to automatically extract hierarchical and sequential features ~\cite{kim2014convolutional, yin2017comparative}. CNNs are efficient at identifying local patterns (e.g., clinical usage phrases), while RNNs and their variants (e.g., DPCNN, RCNN) are better suited for capturing long-range dependencies within sentences ~\cite{johnson2017deep, lai2015recurrent}. Although these models outperform traditional ML in accuracy, they are often considered black boxes and lack native interpretability, necessitating tools like Integrated Gradients or LIME ~\cite{sundararajan2017axiomatic, ribeiro2016lime}.

\subsection{Transformer-Based Models and LLMs}

The introduction of transformer architectures ~\cite{vaswani2017attention} revolutionized NLP by enabling parallel sequence modeling through self-attention mechanisms. Pretrained models such as BERT ~\cite{devlin2018bert} and its Chinese variants ~\cite{cui2019pretraining} have been applied to regulatory tasks by generating rich contextual embeddings for downstream classifiers. More recently, large language models (LLMs) like DeepSeek and LLaMA ~\cite{touvron2023llama, deepseek2024model} have shown potential for zero-shot classification through prompt-based methods.

While LLMs offer unprecedented generalization capabilities, their applicability in regulatory domains remains limited by issues of computational cost, consistency, and lack of transparency ~\cite{banerjee2023large}. Post-hoc interpretability via LIME or token attribution helps address this but introduces challenges related to explanation stability and semantic alignment with human reasoning ~\cite{jacovi2020towards}.

\subsection{Model Interpretability in Regulatory Contexts and Computational Cost}
Interpretability is a critical concern in regulated environments, where model decisions must be traceable and understandable ~\cite{herman2021position, rudin2019stop}. Tools such as SHAP ~\cite{lundberg2017unified}, LIME ~\cite{ribeiro2016lime}, Anchors ~\cite{ribeiro2018anchors}, and Integrated Gradients ~\cite{sundararajan2017axiomatic} have become standard for explaining predictions of black-box models. SHAP in particular enables consistent feature attribution across model types and has been effectively applied in medical AI contexts ~\cite{lundberg2018explainable}. TreeSHAP, a variant tailored for tree ensembles like XGBoost, is especially efficient and faithful ~\cite{lundberg2020local}. However, explanation fidelity tends to degrade with model complexity, particularly for deep or transformer-based models.

Beyond accuracy and interpretability, computational cost is a practical bottleneck in real-world deployment ~\cite{strubell2019energy}. Classical models like Logistic Regression and Naive Bayes offer fast inference and low memory usage, ideal for integration into constrained systems. In contrast, LLMs require significant resources for both inference and training, which can limit their accessibility in clinical and regulatory settings ~\cite{wu2022survey}. Efficient variants of BERT and inference-optimized architectures are being explored to mitigate these barriers ~\cite{sanh2019distilbert}.

\subsection{Trade-off}

Although AI has the potential to enhance regulatory decision-making, model selection requires careful consideration of trade-offs across accuracy, interpretability, and computational feasibility ~\cite{mantus2014fda, pomeranz2013comprehensive}, particularly in high-stakes domains such as healthcare and regulation. Recent work has highlighted that these performance dimensions are often in tension, requiring practitioners to make deliberate design choices based on contextual priorities ~\cite{zhang2024towards}. For instance, while deep learning models tend to achieve superior predictive accuracy, they often do so at the expense of interpretability and computational cost, making them less ideal for settings requiring transparency and auditability. Simpler models like logistic regression or decision trees may provide clearer reasoning pathways but struggle with complex or nuanced inputs. 

Zhang et al.\ proposed a Composite Interpretability (CI) score that quantifies this trade-off by integrating model transparency and explanation fidelity, observing that the relationship between interpretability and performance is not always monotonic and may depend on data domain and model architecture ~\cite{zhang2024towards}. Similarly, studies in the fairness and accountability literature have emphasized that these trade-offs are not merely technical, but are shaped by broader socio-technical constraints, including the epistemic trust of regulators and the institutional requirements for interpretability ~\cite{kaur2022trust}. Chen et al.\ further explore this space by showing how user-facing explanation systems must balance cognitive effort with informativeness, highlighting that interpretability is not a purely objective metric but one that varies with audience expertise and context ~\cite{chen2024taxonomy}.

Our study situates the model selection problem for medical device classification squarely within this trade-off space. By benchmarking traditional ML models, deep learning architectures, and LLMs across accuracy, interpretability, and computational cost, We provide empirically grounded, engineering-oriented guidance for selecting models that align with regulatory objectives—whether the priority is classification accuracy, decision transparency, or scalable deployment.
\section{Methodology}
\begin{figure}[H]
   \centering
\includegraphics[width=0.73\textwidth]{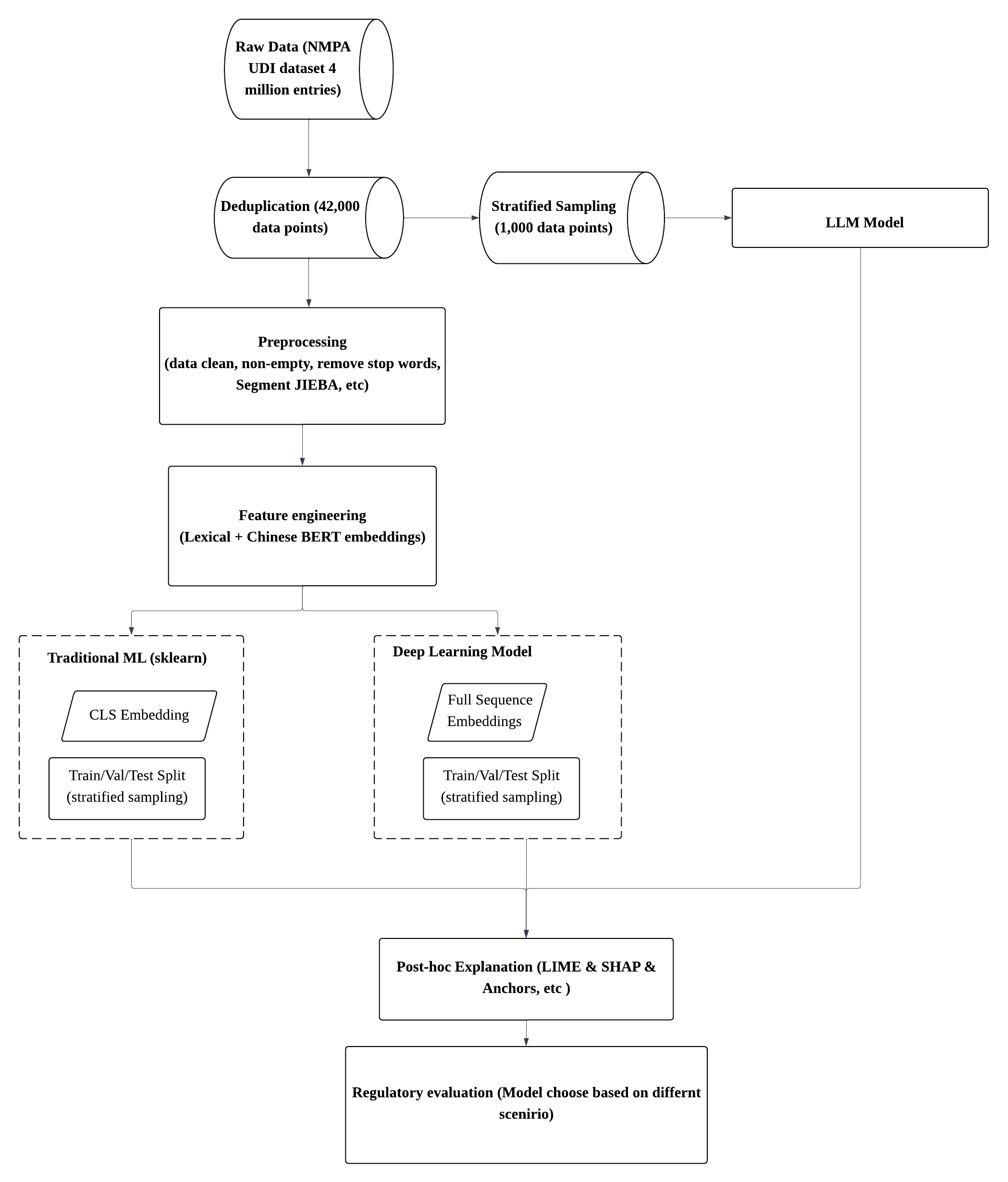}
    \caption{Text process, Model Training and Evaluation Pipeline}
    \label{fig:er}
\end{figure}

In this study, we implement a comprehensive data preprocessing pipeline to prepare medical device descriptions for a classification task. The raw dataset comes from the National Medical Products Administration (NMPA) Unique Device Identification (UDI) database, containing records of medical devices with descriptions in Chinese. We outline the key preprocessing stages below – from data collection and cleaning, through text normalization, embedding generation, and finally splitting the processed data for model training. Each stage is designed to ensure the data is clean, consistent, and properly formatted for input into either traditional machine learning models or deep learning models, as shown in Figure \ref{fig:er}.

\subsection{Dataset and Preprocessing}
We curated a real-world dataset from the China National Medical Products Administration (NMPA) to support the classification of medical devices by regulatory risk. The primary data source was the UDID FULL RELEASE 20240701 registry~\cite{nmpa_udi_database}, which contains over 4 million entries. After deduplication and the removal of incomplete or placeholder records, we retained a high-quality subset of 42,000 labeled samples with complete fields for device description (cpms), product name, and device category. 

Due to the significantly higher computational cost of large language model (LLM) inference, we further sampled 1,000 entries—again using stratified sampling—for our zero-shot classification experiments. The goal of this reduced subset was not to match the predictive performance of fully trained models, but rather to explore the reasoning capabilities and interpretability of LLMs in a constrained yet representative setting. We acknowledge that the use of different dataset sizes introduces some limitations in directly comparing models across experiments. To address this, we clearly distinguish between results from the 42,000-sample experiments and those from the 1,000-sample LLM evaluation, and we caution readers against making performance comparisons without considering these experimental differences.

For classification targets, we used the officially assigned NMPA risk labels—Class I, II, or III—as ground truth. In cases where labels were missing, we inferred the class from the unique identifier field (zczbhhzbapzbh) via character-level parsing. Records with ambiguous or undefined classifications were excluded.

To preserve domain-specific semantic nuances, device descriptions (cpms) were processed in Chinese using pretrained language models. We applied \textsl{jieba} for word segmentation, and input sequences were encoded using the \textsl{hfl/chinese-bert-wwm-ext} model from HuggingFace~\cite{cui2019pretraining}.

Standard preprocessing steps included punctuation removal, stopword filtering, and lexical normalization. Class labels were numerically encoded using \textsl{LabelEncoder} to support multi-class classification. The data was split into training (80\%) and testing (20\%) subsets. To ensure robust evaluation, we applied 10-fold stratified cross-validation to the training set and repeated all experiments across multiple random seeds to assess stability.

\subsection{Text Representation}
We encode the tokenized text descriptions using the bert-base-uncased model. For traditional models (e.g., SVM, Logistic Regression, Random Forest, XGBoost), we extract the 768-dimensional embedding of the [CLS] token as the sentence-level representation, resulting in a feature matrix X\_bert. These models operate on fixed-length vectors and do not retain token-level sequential information.

Deep neural models such as BERT-LSTM and TextCNN utilize the full sequence output from BERT. Instead of summarizing each sentence into a single vector, they take the sequence of contextual token embeddings as input, allowing the model to learn from intra-sentence dependencies and local n-gram structures. Although the preprocessing and embedding stage is consistent across all models, this difference in representation usage allows deep models to capture more fine-grained semantic patterns. The corresponding class labels are encoded numerically with LabelEncoder.

\subsection{Train–Validation–Test Split}
The final step in data preprocessing is to split the prepared dataset into separate sets for model training, validation, and testing. We perform a stratified split to ensure that each subset has the same class distribution as the full dataset. In other words, the proportion of Class I, Class II, and Class III devices is preserved in the training set, the validation set, and the test set (each subset mirrors the overall distribution). This stratification is important for imbalanced or multi-class data to prevent, for example, the test set from having only high-risk devices or an uneven mix. Preserving class proportions across splits leads to a more reliable evaluation – the model’s performance on the test set will reflect performance across all classes, since no class is under- or over-represented by chance~\cite{tasos2019stratified}.

Typically, we use something like an 80/10/10 or 70/15/15 split (e.g., 70\% of the records for training, 15\% for validation, 15\% for testing). The training set is used to train the models (either to fit the scikit-learn classifier or to train/fine-tune the deep neural network). The validation set is held out during training and used to tune hyperparameters and make model selection decisions – for instance, deciding when to stop training to avoid overfitting or which model architecture works best. Finally, the test set is kept completely independent of any training or tuning, and is used at the end to obtain an unbiased evaluation of the model’s classification accuracy on new, unseen data. By using stratified sampling, we ensure that if (for example) Class III devices make up 20\% of the whole dataset, they will be about 20\% of each of the train/val/test sets as well, thus the model is trained and evaluated on representative data from all classes~\cite{tasos2019stratified}.

\subsection{Classifiers and Training Setup}

To evaluate the effectiveness of different modeling strategies for regulatory medical device classification, we implement and compare three families of classifiers: traditional machine learning models, deep neural network architectures, and large language models (LLMs). All models are trained and assessed using 10-fold stratified cross-validation, and performance is evaluated on a held-out test set using standard metrics including accuracy, precision, recall, and F1-score. We also log training and inference times to assess computational efficiency.

For input representation, we use the Chinese pre-trained BERT model \textsl{hfl/chinese-bert-wwm-ext} from HuggingFace~\cite{cui2019pretraining}. Device descriptions are tokenized and encoded using this model, with the final [CLS] token embedding extracted to form a 768-dimensional fixed-length input vector.

\paragraph{Traditional Machine Learning Models.} We evaluate four widely used classifiers: Logistic Regression (LR), Gaussian Na\"ive Bayes (GNB), Support Vector Machine (SVM), Random Forest (RF), and XGBoost (XGB). LR is configured with L2 regularization and class-balanced weights. SVM uses an RBF kernel with hyperparameters tuned via GridSearchCV over $C \in \{0.1, 1, 10\}$. RF includes 100 decision trees trained with class weights, while XGB is configured for multi-class classification using \textsl{mlogloss}, with tree depth $\in \{4, 6, 8\}$ and learning rate $\in \{0.05, 0.1\}$. All models use BERT [CLS] embeddings as input and are implemented in scikit-learn~\cite{scikit-learn}.

\paragraph{Deep Learning Models.} We implement Convolutional Neural Networks (CNN), Deep Pyramid CNN (DPCNN), Recurrent Neural Networks (RNN), and Long Short-Term Memory (LSTM) models using TensorFlow/Keras. All models take input shaped \texttt{(n, 768, 1)} from BERT embeddings.

The CNN model consists of a 1D convolutional layer with 128 filters and kernel size 3 (\textsl{padding='same'}), followed by \textsl{GlobalMaxPooling1D}, a dropout layer (rate = 0.5), a dense layer with 64 ReLU units, and a softmax output layer. The LSTM model includes a single unidirectional LSTM with 128 units, followed by dropout (0.5), a dense layer (64 ReLU), and a softmax layer. Both models are compiled with the Adam optimizer (learning rate = 0.001), \textsl{sparse\_categorical\_crossentropy} loss, batch size of 32, 10 epochs, and a 20\% validation split.

\paragraph{Large Language Models (LLMs).} To explore zero-shot classification, we employ \textsl{DeepSeek-R1-Distill-Qwen-7B}
via the Ollama local API. We use a 10-sample subset of our dataset, each labeled with a regulatory class inferred from the NMPA identifier field (\textsl{zczbhhzbapzbh column}). 

Each prompt simulates expert reasoning using NMPA risk classification:

\begin{quote}
\small
\textit{You are a senior medical device regulatory affairs specialist familiar with the NMPA (National Medical Products Administration) classification system in China. Please classify the following medical device based on its description into one of the three NMPA regulatory risk classes:} \\
\textbf{Class I} (Low Risk) \\
\textbf{Class II} (Medium Risk) \\
\textbf{Class III} (High Risk) \\
\textbf{Device Description:} \textsl{\{text\}} \\
\textbf{Your answer must be ONLY one of the following:} Class I, Class II, or Class III.
\end{quote}

The LLM is asked to respond with a single class label based on its internal reasoning and the structured prompt. This setup enables evaluation of whether LLMs can simulate regulatory logic even in zero-shot settings, and how their interpretability compares to other models.

All models were evaluated for classification metrics as well as runtime and memory usage using the \textsl{psutil} library. These experiments allow for direct comparison across model types in terms of practical deployment within a regulatory workflow.

\subsection{Interpretability}
Interpretability is a foundational requirement in regulated domains such as medical device classification, where transparency, auditability, and trust are not optional but mandated by law. In this study, we systematically evaluate the interpretability of each model—traditional machine learning, deep neural networks, and large language models (LLMs)—using methods appropriate to their structural characteristics. Our goal is to compare models not only on their predictive performance but also on how clearly and efficiently their decisions can be understood by domain experts.

The models are categorized into two broad classes: intrinsically interpretable models (e.g., Logistic Regression, Naïve Bayes), and black-box models requiring post-hoc explanation (e.g., SVM, XGBoost, CNNs, and LLMs) ~\cite{rudin2019stop}. For intrinsically interpretable models, we leverage native explanations: Logistic Regression coefficients directly represent global feature importance across classes, while Naïve Bayes allows inspection of class-conditional likelihoods $P(x_i \mid y)$. To enhance local interpretability, we supplement with LIME and SHAP visualizations ~\cite{ribeiro2016lime, lundberg2017unified}.

For SVMs, which lack transparency due to kernelization, we use LIME and KernelSHAP to approximate local feature attributions through perturbation-based sampling~\cite{lundberg2017unified}. Permutation-based global importance is also computed to aid summary-level interpretation. For XGBoost, we adopt TreeSHAP~\cite{lundberg2020local}, which provides exact and efficient Shapley values tailored for tree ensembles, enabling interpretable deployment in production settings.

For deep learning models such as CNNs, RNNs, and DPCNNs, we apply input-specific post-hoc techniques: Integrated Gradients~\cite{sundararajan2017axiomatic} and LIME-Text for token-level attribution~\cite{ribeiro2016lime}. These methods reveal which parts of the input text contribute most to the model's decision.

For LLMs (e.g., DeepSeek, LLaMA), we employ token-level LIME. Although SHAP becomes infeasible at this scale, LIME offers a lightweight approximation of feature influence. We also visualize attention maps where relevant~\cite{jacovi2020towards}, though we acknowledge their limitations in reliably reflecting causal importance. These transformer-based architectures present unique interpretability challenges: LIME requires significant sampling, explanations are sensitive to perturbation, and attribution outputs (e.g., attention scores or token masks) may not align with human semantic understanding. These issues are critical in regulatory applications where explanations must be stable, auditable, and human-comprehensible.

\begin{table}[H]
\centering
\caption{Interpretability Strategy by Model Type and Input Format}
\begin{tabular}{|l|l|}
\hline
\textbf{Model Type} & Na\"ive Bayes \\
\hline
\textbf{Input Format} & TF-IDF \\
\textbf{Primary Method} & Direct inspection of \textsl{feature\_log\_prob\_} \\
 \\
\textbf{Supplementary Tools} & LIME, SHAP \\
\hline
\textbf{Model Type} & Logistic Regression \\
\hline
\textbf{Input Format} & TF-IDF or BERT embeddings \\
\textbf{Primary Method} & Anchors, Coefficients \\
\textbf{Supplementary Tools} & LIME, SHAP \\
\hline
\textbf{Model Type} & SVM \\
\hline
\textbf{Input Format} & Raw text \\
\textbf{Primary Method} & LIME-Text \\
\textbf{Supplementary Tools} & Anchors, KernelSHAP, Permutation \\
\hline
\textbf{Model Type} & XGBoost / Random Forest \\
\hline
\textbf{Input Format} & BERT embeddings  \\
\textbf{Primary Method} & TreeSHAP \\
\textbf{Supplementary Tools} & Permutation Importance \\
\hline
\textbf{Model Type} & CNN / RNN / DPCNN \\
\hline
\textbf{Input Format} & Raw tokenized text \\
\textbf{Primary Method} & Integrated Gradients \\
\textbf{Supplementary Tools} & LIME-Text \\
\hline
\end{tabular}
\label{tab:interpretability_methods_vertical}
\end{table}

We match input formats to model capabilities for optimal performance and interpretability. Traditional models like Logistic Regression and Naïve Bayes operate effectively on vectorized representations (e.g., TF-IDF or BERT embeddings), which preserve global lexical patterns and are interpretable through coefficient-based weights~\cite{ribeiro2016lime}. Tree-based models (XGBoost, Random Forest) benefit from high-dimensional embeddings like BERT due to their ability to capture non-linear interactions without explicit sequence modeling~\cite{lundberg2020local}. For neural architectures (CNNs, RNNs, LLMs), raw or tokenized text inputs are preserved to leverage their capacity for sequential feature learning and context modeling. Particularly for SVMs and LLMs, using raw text enables token-level attribution via LIME-Text or attention-based methods, which are more faithful to the linguistic structure of the input~\cite{jacovi2020towards}.

\section{Experiments and Results}
\label{sec:experiments}

We evaluate all models across three dimensions: predictive accuracy, interpretability, and computational cost. These dimensions are essential in the regulatory context, where decisions must be both reliable and explainable, while also being efficient for deployment.

\subsection{Model Accuracy}

To assess classification performance, we compute the following metrics using 10-fold stratified cross-validation on 42,000 data points:
\begin{itemize}
    \item \textbf{Weighted F1 Score}: Measures the harmonic mean of precision and recall, weighted by support.
    \item \textbf{Weighted Precision and Recall}: Captures the balance between false positives and false negatives across classes.
    \item \textbf{Area Under ROC Curve (AUROC)}: One-vs-Rest strategy for multi-class evaluation.
    \item \textbf{Accuracy}: The proportion of correctly predicted instances.
\end{itemize}

Based on overall accuracy as shown in Table \ref{tab:ml_performance} and Table \ref{tab:dl_performance}, the CNN model outperformed all other approaches, achieving an accuracy of 88.28\%, followed by XGBoost (86.00\%) and Random Forest (84.00\%). Among traditional machine learning models, XGBoost demonstrated the most balanced performance across metrics, with a Macro F1 of 0.69 and strong performance in high-risk Class III devices (F1 = 0.87). Logistic Regression remained the most computationally efficient (0.0208s) while maintaining reasonably competitive performance (Accuracy = 82.00\%, Macro F1 = 0.62), making it attractive for scenarios where transparency and low latency are prioritized.

Within deep learning, CNN clearly outperformed RCNN, RNN, and DPCNN, achieving the highest overall F1 (0.6768) and superior Class III performance (F1 = 0.80). Other architectures struggled to generalize as well, particularly on Class I devices, with RNN and DPCNN showing Class I F1 scores below 0.15, reinforcing that CNNs are better suited for localized textual pattern recognition in regulatory descriptions.

In contrast, large language models (LLMs) performed substantially worse on accuracy (DeepSeek R1: 45.93\%), despite showing moderate recall for Class II and III devices (around 60--70\%) as shown in Table \ref{tab:llm_performance}. This suggests that zero-shot LLMs can approximate regulatory heuristics, but without domain-specific tuning, they fall short in precision and struggle with low-risk class identification (Class I F1 < 0.15).

Overall, while LLMs provide human-readable rationales, their performance trade-offs remain substantial. In practical deployments, models such as XGBoost and CNN currently offer the best balance between accuracy, class-level reliability, interpretability (via SHAP or Integrated Gradients), and computational efficiency.

\begin{table}[H]
\centering
\caption{Performance of Traditional Machine Learning Models}
\label{tab:ml_performance}
\resizebox{\textwidth}{!}{%
\begin{tabular}{|l|c|c|c|c|c|}
\hline
\textbf{Model} & \textbf{Accuracy} & \textbf{Macro F1} & \textbf{Macro Precision} & \textbf{Macro Recall} & \textbf{Inference Time (s)} \\
\hline
SVM & 0.84 & 0.67 & 0.65 & 0.69 & 82.87 \\
Logistic Regression & 0.82 & 0.62 & 0.60 & 0.67 & 0.0208 \\
Random Forest & 0.84 & 0.67 & 0.76 & 0.63 & 0.2272 \\
XGBoost & 0.86 & 0.69 & 0.74 & 0.66 & 0.0514 \\
Naive Bayes & 0.52 & 0.40 & 0.45 & 0.51 & 0.1371 \\
\hline
\end{tabular}
}

\vspace{1em}

\resizebox{\textwidth}{!}{%
\begin{tabular}{|l|ccc|ccc|}
\hline
\textbf{Model} & \textbf{Class I F1} & \textbf{Class II F1} & \textbf{Class III F1} & \textbf{Class I Recall} & \textbf{Class II Recall} & \textbf{Class III Recall} \\
\hline
SVM & 0.31 & 0.84 & 0.85 & 0.39 & 0.83 & 0.85 \\
Logistic Regression & 0.20 & 0.81 & 0.83 & 0.37 & 0.80 & 0.83 \\
Random Forest & 0.32 & 0.84 & 0.84 & 0.22 & 0.86 & 0.82 \\
XGBoost & 0.34 & 0.85 & 0.87 & 0.26 & 0.85 & 0.87 \\
Naive Bayes & 0.03 & 0.56 & 0.61 & 0.50 & 0.50 & 0.53 \\
\hline
\end{tabular}
}
\end{table}

\begin{table}[H]
\centering
\caption{Performance of Deep Learning Models}
\label{tab:dl_performance}
\resizebox{\textwidth}{!}{%
\begin{tabular}{|l|c|c|c|c|c|}
\hline
\textbf{Model} & \textbf{Accuracy} & \textbf{Macro F1} & \textbf{Macro Precision} & \textbf{Macro Recall} & \textbf{Inference Time (s)} \\
\hline
CNN & 0.8828 & 0.6768 & 0.7102 & 0.6573 & 0.0034 \\
RCNN & 0.7550 & 0.5047 & 0.5224 & 0.5083 & 0.0034 \\
DPCNN & 0.6574 & 0.4370 & 0.5011 & 0.4435 & 0.0024 \\
RNN & 0.7384 & 0.4938 & 0.5044 & 0.4956 & 0.0031 \\
\hline
\end{tabular}
}

\vspace{1em}

\resizebox{\textwidth}{!}{%
\begin{tabular}{|l|ccc|ccc|}
\hline
\textbf{Model} & \textbf{Class I F1} & \textbf{Class II F1} & \textbf{Class III F1} & \textbf{Class I Recall} & \textbf{Class II Recall} & \textbf{Class III Recall} \\
\hline
CNN & 0.50 & 0.73 & 0.80 & 0.39 & 0.75 & 0.83 \\
RCNN & 0.20 & 0.59 & 0.72 & 0.19 & 0.61 & 0.73 \\
DPCNN & 0.12 & 0.42 & 0.75 & 0.14 & 0.38 & 0.80 \\
RNN & 0.14 & 0.52 & 0.76 & 0.12 & 0.52 & 0.80 \\
\hline
\end{tabular}
}
\end{table}

\begin{table}[H]
\centering
\caption{Performance of Large Language Models (LLMs) on 1000 Samples}
\label{tab:llm_performance}
\resizebox{\textwidth}{!}{%
\begin{tabular}{|l|c|c|c|c|}
\hline
\textbf{Model} & \textbf{Accuracy} & \textbf{Macro F1} & \textbf{Macro Recall} & \textbf{Inference Time (s)} \\
\hline
DeepSeek R1 7B & 0.4593 & 0.2534 & 0.3854 & 3.9888 \\
LLaMA 3.1 8B & 0.4553 & 0.2498 & 0.2919 & 0.4919 \\
\hline
\end{tabular}
}

\vspace{1em}

\resizebox{\textwidth}{!}{%
\begin{tabular}{|l|ccc|ccc|}
\hline
\textbf{Model} & \textbf{Class I F1} & \textbf{Class II F1} & \textbf{Class III F1} & \textbf{Class I Recall} & \textbf{Class II Recall} & \textbf{Class III Recall} \\
\hline
DeepSeek R1 7B & 0.15 & 0.58 & 0.57 & 0.14 & 0.60 & 0.72 \\
LLaMA 3.1 8B & 0.13 & 0.56 & 0.57 & 0.12 & 0.55 & 0.70 \\
\hline
\end{tabular}
}
\end{table}

\subsection{Model Interpretability}
In our interpretability experiments, we applied a variety of methods to evaluate different model types and their interpretability mechanisms. Intrinsic models (e.g., logistic regression, naive Bayes) provided stable and globally coherent explanations by design. Feature coefficients and class-conditional likelihoods were directly accessible and aligned with domain expectations, requiring minimal computational overhead. For Naïve Bayes, which uses TF-IDF as the input format, we employed direct inspection of feature log probabilities as the primary method. The results are in Table \ref{tab:log_probabilities}. The log probability tells us the likelihood of a feature appearing in a particular class. A more negative value indicates a less likely occurrence for that feature in that class. For example, "cutting" with a log probability of -7.7423 in Class I suggests that this term is relatively common for Class I devices, while features with less negative log probabilities are more closely tied to the class.

\begin{table}[H]
\centering
\begin{tabular}{|c|l|c|}
\hline
\textbf{Class} & \textbf{Feature} & \textbf{Log Probability} \\
\hline
Class I & cutting & -7.7423 \\
Class I & parts & -7.8036 \\
Class I & orthodontic & -7.8257 \\
Class I & laminoscopes & -7.8427 \\
Class I & staining & -7.8553 \\
Class I & usually & -7.8912 \\
Class I & transfer & -7.8912 \\
Class I & linear & -7.8937 \\
Class I & components & -7.9106 \\
Class I & non & -7.9106 \\
\hline
Class II & disposable & -5.8229 \\
Class II & consists & -5.9371 \\
Class II & product & -5.9375 \\
Class II & used & -5.9843 \\
Class II & tube & -6.0703 \\
Class II & stapler & -6.1272 \\
Class II & medical & -6.1384 \\
Class II & needle & -6.2938 \\
Class II & body & -6.3645 \\
Class II & type & -6.3737 \\
\hline
Class III & product & -5.6874 \\
Class III & needle & -6.0693 \\
Class III & catheter & -6.1376 \\
Class III & screws & -6.1545 \\
Class III & infusion & -6.3185 \\
Class III & consists & -6.3410 \\
Class III & used & -6.3796 \\
Class III & bone & -6.4495 \\
Class III & sterilized & -6.4546 \\
Class III & tube & -6.4946 \\
\hline
\end{tabular}
\caption{Naïve Bayes Model: Log Probabilities for Features Across Different Classes}
\label{tab:log_probabilities}
\end{table}

\begin{figure}[H]
    \centering
    \begin{subfigure}[t]{1\textwidth}
        \centering
        \includegraphics[width=\linewidth]{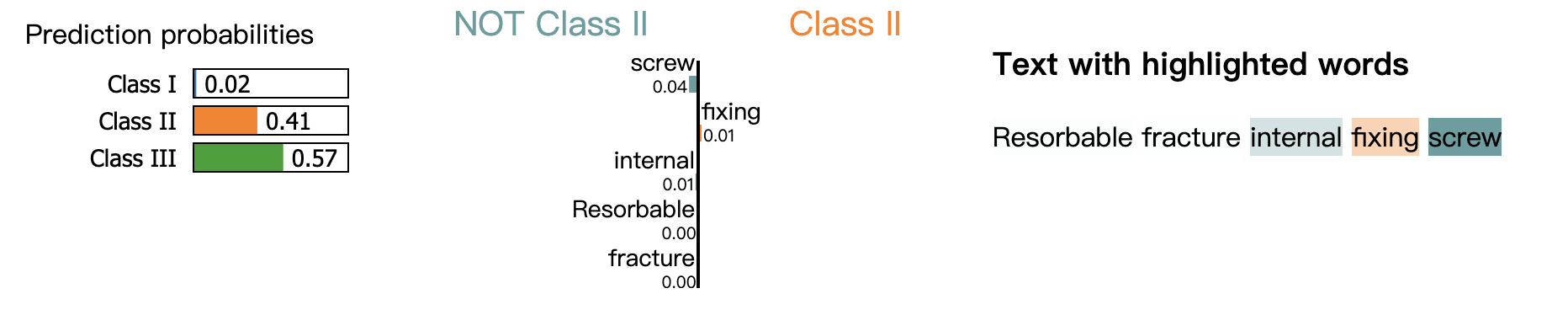}
        \caption{SVM LIME Explanation}
        \label{fig:svm_lime}
    \end{subfigure}
    
    \vspace{0.5cm}
    
    \begin{subfigure}[t]{1\textwidth}
        \centering
        \includegraphics[width=\linewidth]{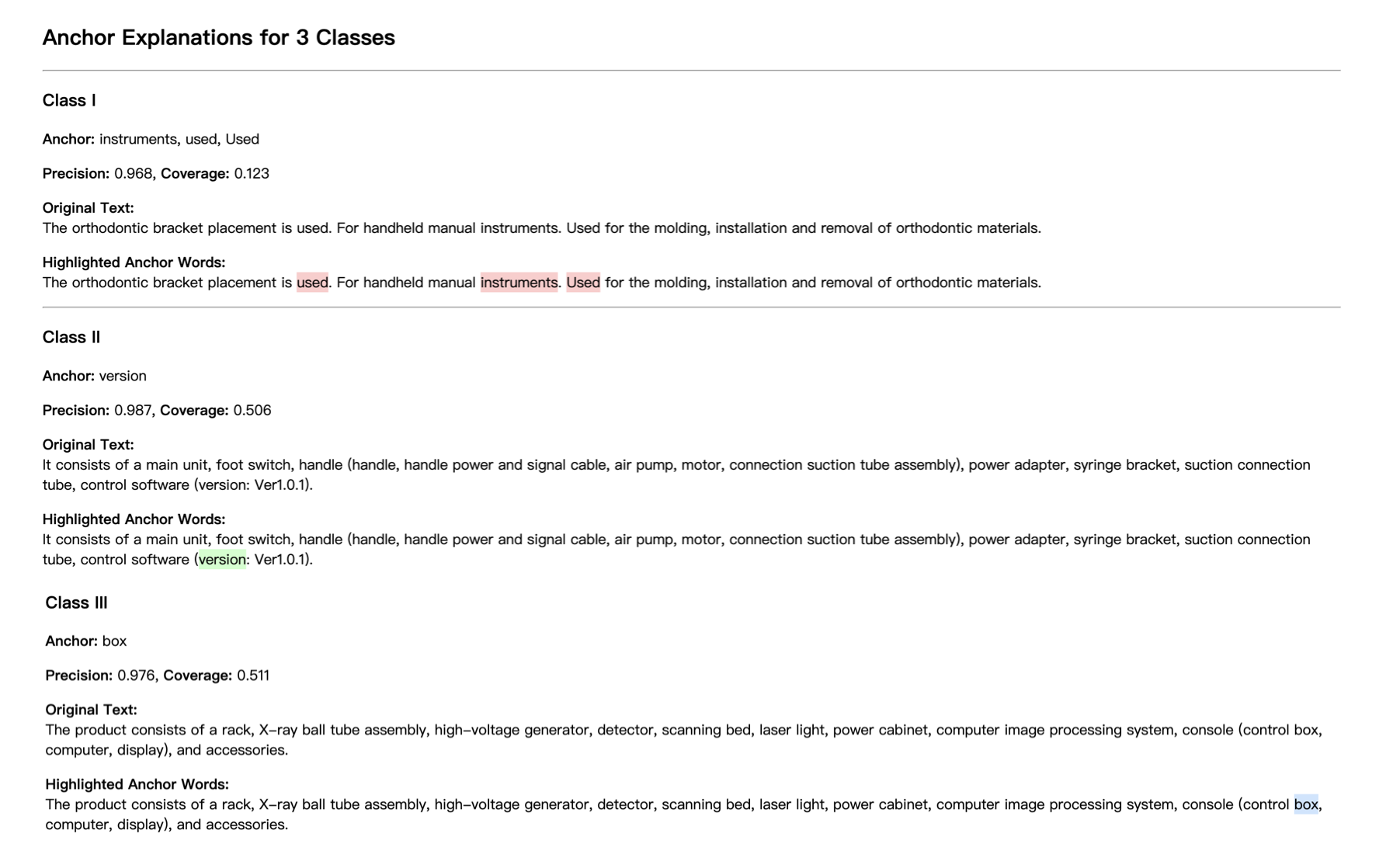}
        \caption{Logistic Regression Anchor}
        \label{fig:logreg_anchor}
    \end{subfigure}
    
    \vspace{0.5cm}
    
    \begin{subfigure}[t]{1\textwidth}
        \centering
        \includegraphics[width=\linewidth]{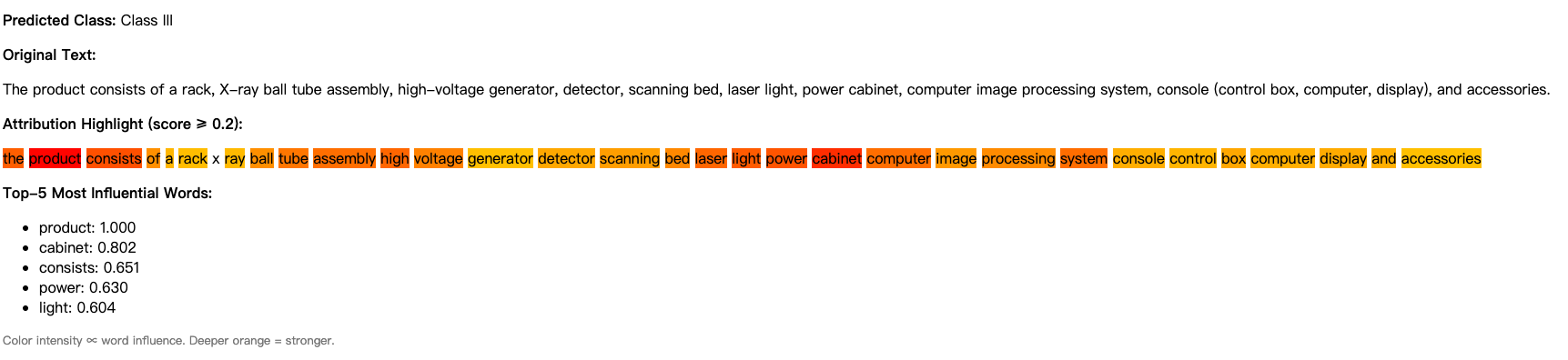}
        \caption{RNN Integrated Gradients Explanation}
        \label{fig:rnn_ig}
    \end{subfigure}
    
    \caption{Comparative interpretability visualizations across models: (a) SVM with LIME, (b) Logistic Regression with Anchor explanations, and (c) RNN with Integrated Gradients.}
    \label{fig:combined_interpretability}
\end{figure}

Logistic Regression, which utilizes either TF-IDF or BERT embeddings, was examined through Anchors as shown in Figure \ref{fig:combined_interpretability}. For Class I, the anchor words identified were “instruments” and “used”, with a high precision (96.8\%) but low coverage (12.3\%), suggesting that while these words are reliable indicators of low-risk instruments, they appear only in a limited subset of Class I samples. For Class II, the anchor word “version” exhibited both high precision (98.7\%) and substantial coverage (50.6\%), indicating that references to software versions or firmware updates are not only distinctive but also prevalent among Class II devices. For Class III, the model consistently relied on the anchor word “box” (e.g., control box), with a precision of 97.5\% and coverage of 51.1\%, reflecting the common presence of control hardware in high-risk, complex systems. This comparative analysis highlights how the model’s interpretability varies by class: Class I relies on sparse but specific cues, while Class II and III benefit from broader, high-impact lexical anchors. These findings reinforce the importance of class-specific vocabulary in medical device classification and support the model's use in regulatory decision-making contexts.

For Support Vector Machines (SVM), we can work with just raw text input, focusing on LIME-Text as the primary interpretability tool. As shown in Figure \ref{fig:combined_interpretability}, the text is analyzed by highlighting the most influential words, which are identified based on their contribution to the predicted class. In this specific example, the model predicts the probabilities for three possible classes: Class I, Class II, and Class III. The prediction probabilities suggest that Class III has the highest likelihood (0.57), followed by Class II (0.41), and Class I (0.02). The words "Resorbable", "fracture", "internal", "fixing", and "screw" are highlighted according to their significance in the decision-making process. In the context of Class II, words like "fixing" have higher weights, which influence the model's prediction that this sample belongs to Class II. However, the word "screw" is more strongly associated with Class III, indicating that it contributes to a higher probability for Class III. By using LIME-Text, we can visually explain how each word affects the model's output and interpret which parts of the input are driving the classification, helping to make the SVM's decision-making process more transparent and interpretable.

In the case of tree-based models like XGBoost and Random Forest, we used BERT embeddings and TreeSHAP as the primary method. When using BERT embeddings, we visualize the embedding vectors rather than specific words. This adds an abstraction layer to the explanation, as BERT embeddings represent a more complex, high-dimensional space that captures the semantic relationships between words. Therefore, when visualizing BERT embeddings, we interpret the relationships between these vector representations, which may not directly correspond to individual word-level contributions. This distinction impacts how we interpret the importance of different parts of the text, as we are not directly attributing the importance to individual words but rather to the abstracted embedding representations.

Explanations for black-box models such as SVM, XGBoost, and deep neural networks depended heavily on the quality of local approximations. Although LIME successfully generated sparse, human-readable explanations for individual predictions across all models, its stability varied. For example, SVM explanations were occasionally sensitive to sampling randomness due to decision boundary complexity. In the case of XGBoost, TreeSHAP yielded highly stable and faithful explanations, often outperforming LIME in both consistency and global insight.

Although CNNs and RNNs are considered "black-box" models, they still process raw tokenized text (i.e., inputting text directly into the model after tokenization), and Integrated Gradients is used to explain their predictions. Integrated Gradients helps to understand the importance of individual tokens (words) by attributing the prediction to each token based on how much it affects the output when moving from a baseline to the actual input. This method is particularly effective for deep learning models that process raw text because it works well with the continuous nature of deep models, offering insights into how each word influences the model’s decision as shown in Figure \ref{fig:combined_interpretability}.

\begin{figure}[H]
   \centering
    \includegraphics[width=0.5\textwidth]{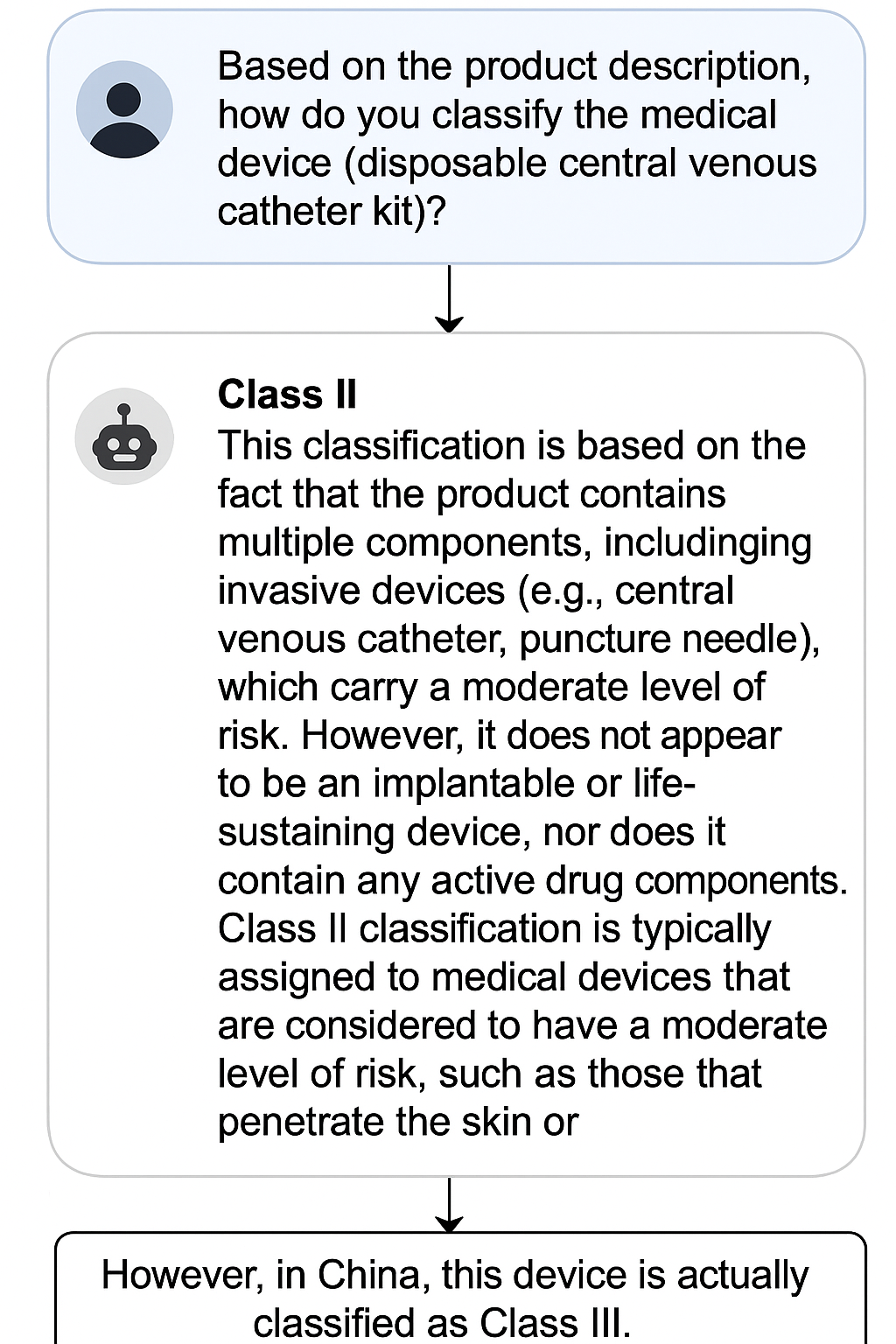}
    \caption{LLM interpretability}
    \label{fig:llmm}
\end{figure}

Finally, LLMs exhibited the greatest interpretability for understanding why the model predicts that a certain device belongs to a particular classification. Due to their ability to process and generate natural language, LLMs provide intuitive, human-readable explanations, making the reasoning behind their decisions more transparent. They are able to explain why a device might belong to a certain class based on contextual understanding and domain knowledge, making it easier to trace the logic of their predictions as shown in Figure \ref{fig:llmm}. However, despite the superior interpretability of LLMs, the accuracy results indicate a different story. We observed that the accuracy rate of the LLM-based model was relatively low compared to other models like SVM or XGBoost as shown in Table \ref{tab:llm_performance}. This suggests that while LLMs aim to provide insightful and explainable reasoning for their decisions, they do not always offer the correct legal reasoning.

\begin{figure}[H]
    \centering
    \begin{subfigure}[t]{0.48\textwidth}
        \centering
        \includegraphics[width=\linewidth]{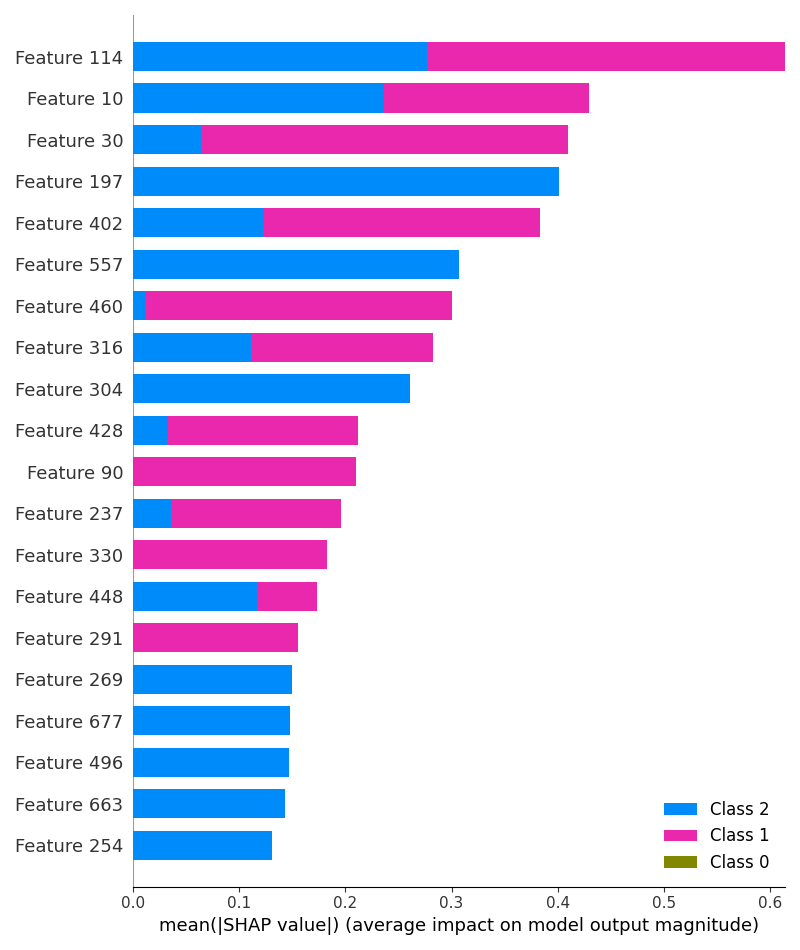}
        \caption{Random Forest BERT TreeSHAP}
        \label{fig:rf_treeshap}
    \end{subfigure}
    \hfill
    \begin{subfigure}[t]{0.48\textwidth}
        \centering
        \includegraphics[width=\linewidth]{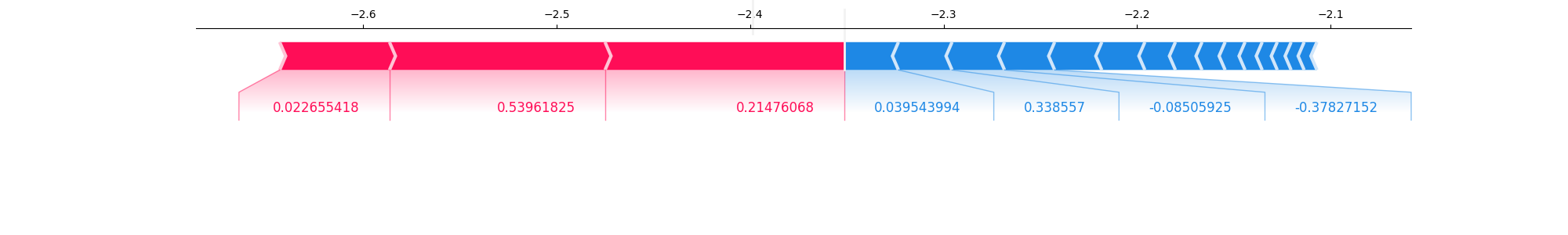}
        \caption{XGBoost BERT TreeSHAP}
        \label{fig:xgb_treeshap}
    \end{subfigure}
    \caption{TreeSHAP-based explanations for BERT embeddings using ensemble models.}
    \label{fig:treeshap_comparison}
\end{figure}

While interpretability is often cited as a key requirement in medical regulatory AI systems, there exists no universally accepted benchmark or ground truth to evaluate the quality of explanations. We benchmarked SHAP, LIME, Anchors, and Integrated Gradients across multiple models (Logistic Regression, SVM, XGBoost, etc.), yet acknowledge that explanation quality is inherently model- and context-dependent. In regulatory practice, the primary goal is not necessarily to achieve the most accurate explanation, but to ensure that model decisions align with domain knowledge and are auditable. 

Evaluating interpretability remains an open challenge, particularly in regulated AI systems where no ground-truth explanations exist. Rather than claiming superiority of a single method, we adopt a multi-method strategy to triangulate model behavior.

While models such as XGBoost and Random Forest demonstrate strong performance when operating on BERT embeddings, their interpretability remains limited due to the nature of these inputs. Specifically, methods like TreeSHAP identify the contribution of individual embedding dimensions to model predictions, but these dimensions do not correspond directly to human-readable words or phrases. This introduces an abstraction layer between the model's internal logic and the regulator’s semantic understanding of why a prediction was made. This gap poses real challenges for regulatory auditability and decision traceability. Regulatory professionals typically require transparent, token-level justifications grounded in recognizable medical or functional terminology. However, the latent representations learned by deep models—while powerful in capturing contextual nuances—lack this transparency. To address this, we propose hybrid interpretability strategies. For instance, embedding-based models can be complemented by manually engineered features that capture domain-relevant tokens (e.g., risk indicators or functional keywords), enabling parallel inspection. Alternatively, simplified proxy models (e.g., rule-based systems or logistic regression trained on key features) can be used alongside embedding models to triangulate decisions and provide more interpretable justifications for end-users. This hybrid approach may help balance the trade-off between representational richness and semantic clarity in high-stakes regulatory contexts.

\subsection{Model Computational Cost}

\begin{table}[ht]
\centering
\resizebox{\textwidth}{!}{
\begin{tabular}{|l|l|l|l|}
\hline
\textbf{Model}              & \textbf{Theoretical Complexity (Big O)}  & \textbf{Observed Inference Time (s)} & \textbf{Inference Time Comments} \\ \hline
Logistic Regression         & $O(nd)$                                  & 0.0208                               & Extremely efficient in resource-constrained environments. \\ \hline
Naive Bayes                 & $O(nd)$                                  & 0.1371                               & Efficient with low inference time. \\ \hline
SVM                         & $O(n^2)$ to $O(n^3)$                      & 82.8700                              & High latency due to kernel computation, not practical for real-time use. \\ \hline
Random Forest               & $O(k \log n)$                            & 0.2272                               & Fast inference with sublinear complexity for tree-based models. \\ \hline
XGBoost                     & $O(k \log n)$                            & 0.0514                               & Efficient, balanced performance and efficiency. \\ \hline
CNN                         & $O(n \cdot d \cdot L)$                   & 0.0024                               & Fast inference, high accuracy, suitable for practical deployment. \\ \hline
RNN                         & $O(n \cdot d \cdot L)$                   & 0.0034                               & Fast inference, suitable for real-time applications. \\ \hline
DeepSeek R1                 & $O(n^2)$                                 & 3.9888                               & High inference time, not feasible for real-time regulatory tasks. \\ \hline
LLaMA                       & $O(n^2)$                                 & 0.4919                               & Relatively high latency, may be impractical for real-time systems. \\ \hline
\end{tabular}
}
\caption{Comparison of Theoretical Complexity and Observed Inference Time for Different Models}
\label{tab:ml_comparison}
\end{table}

We analyze computational cost by linking the theoretical complexity of each model with its observed inference time in practice. While theoretical analysis offers a general understanding of algorithmic efficiency, inference time provides an empirical measure of how these costs manifest in real-world medical classification tasks. Overall, the trend in inference time aligns with the Big O complexity.

\paragraph{Theoretical Complexity.} Classical models such as Logistic Regression and Naive Bayes have linear time complexity $O(nd)$ for training and inference, where $n$ is the number of samples and $d$ is the number of features. In contrast, kernel-based Support Vector Machines (SVMs) scale poorly with training data, exhibiting worst-case complexity of $O(n^3)$ during training. Random Forests and XGBoost ensembles reduce complexity by parallelizing tree operations, yet their inference cost scales with the number of estimators. Deep learning models such as CNNs and RNNs require $O(n \cdot d \cdot L)$ operations for each input, with $L$ denoting network depth. Finally, Large Language Models (LLMs) like DeepSeek and LLaMA incur substantially higher cost, both due to their parameter scale (billions of weights) and prompt-based inference mechanisms.

\paragraph{Empirical Evaluation.} Inference times were measured for all models on the same hardware using 1000 medical device samples. As shown in Tables~\ref{tab:ml_performance}–\ref{tab:dl_performance}, classical models demonstrated superior efficiency. Logistic Regression achieved sub-second inference ($0.0208 \pm 0.0008$ s) while maintaining competitive accuracy. XGBoost offered a favorable balance ($0.0514 \pm 0.0034$ s) between performance and efficiency, owing to optimized tree traversal. Conversely, SVM suffered from extremely high latency ($82.87 \pm 0.59$ s), rendering it impractical in real-time scenarios despite strong accuracy.

Deep neural models exhibited low inference latency across the board (ranging from 0.0024 to 0.0034 s), with CNNs achieving both the highest accuracy (0.8828) and fast response times. These models offer a desirable trade-off for practical deployment, particularly when interpretability is enhanced through methods like Integrated Gradients.

LLMs, while impressive in zero-shot generalization, revealed orders-of-magnitude slower inference. DeepSeek R1 required nearly 4 seconds per input (3.9888 s), and even optimized variants like LLaMA 3.1 consumed 0.4919 s per sample. This latency, combined with memory demands, poses significant barriers for integration into high-throughput regulatory pipelines.

Since interpretability are no strict numerical benchmark, cost is also case by case, we summarize each model accuracy, interpretability and cost by Numerical Rating Criteria for Accuracy, Interpretability, and Computational Cost in Table \ref{tab:criteria}.

\scriptsize
\begin{center}
\textbf{Table 1: Comparison of AI Models in Terms of Accuracy, Interpretability, and Computational Cost}
\end{center}

\begin{tabular}{|p{2.0cm}|p{1.6cm}|p{1.6cm}|p{1.6cm}|p{6.2cm}|}
\hline
\textbf{Model} & \textbf{Accuracy} \newline \textbf{(Stars)} & \textbf{Interpretability} \newline \textbf{(Stars)} & \textbf{Computational Cost} \newline \textbf{(Stars)} & \textbf{Application Scenarios} \\
\hline
Rule-Based & 5 & 5 & 5 & Ideal for regulatory audits, legal compliance, and highly interpretable tasks where decision transparency is critical. Suitable for environments where accuracy is less critical but Interpretability and low computational cost are essential. \\
\hline
Logistic Regression & 4 & 3 & 3 & Suitable for transparent, low-complexity predictions with a good balance of accuracy and Interpretability. Often used in settings where interpretability is more important than maximum accuracy, such as in business analytics or healthcare where decisions need to be explained. \\
\hline
SVM & 5 & 3 & 5 & Best for high-accuracy classification tasks that require moderate interpretability. Often used in domains like image recognition or high-dimensional data where predictive performance is a priority, but the computational cost may be higher. \\
\hline
Random Forest & 5 & 1 & 3 & Ideal for tasks requiring high accuracy with balanced performance and interpretability, particularly for tabular data. However, the lack of Interpretability makes it less suited for regulatory applications where transparency is required. \\
\hline
XGBoost & 5 & 1 & 3 & Perfect for tasks involving complex structured data and competitions. High accuracy but moderate Interpretability, suitable for industrial tasks and machine learning competitions, where performance is paramount but transparency can be sacrificed. \\
\hline
Gaussian Naive Bayes & 1 & 3 & 3 & Simple baseline models for quick prototyping or initial benchmarks. Lower accuracy but with decent Interpretability, often used in environments where rapid model development is needed, such as exploratory data analysis or educational contexts. \\
\hline
\end{tabular}

\pagebreak

\scriptsize
\begin{center}
\textbf{Table 1 (continued): Comparison of AI Models in Terms of Accuracy, Interpretability, and Computational Cost}
\end{center}

\begin{tabular}{|p{2.0cm}|p{1.6cm}|p{1.6cm}|p{1.6cm}|p{6.2cm}|}
\hline
\textbf{Model} & \textbf{Accuracy} \newline \textbf{(Stars)} & \textbf{Interpretability} \newline \textbf{(Stars)} & \textbf{Computational Cost} \newline \textbf{(Stars)} & \textbf{Application Scenarios} \\
\hline
DPCNN & 3 & 3 & 1 & Best for complex text classification tasks with a balance between accuracy and Interpretability. Ideal for applications in natural language processing where high accuracy is necessary but low computational cost is required. \\
\hline
CNN & 5 & 3 & 1 & Well-suited for tasks in text and sequence pattern recognition, including image and speech recognition. High accuracy with moderate Interpretability, making it ideal for practical deployment where computational resources are less constrained. \\
\hline
RCNN & 4 & 3 & 1 & Good for sequence labeling and moderate complexity tasks. Often used in NLP applications requiring a trade-off between accuracy, Interpretability, and computational efficiency. \\
\hline
RNN & 4 & 3 & 1 & Ideal for sequential data and NLP applications, where the ability to process sequences is critical. Moderate Interpretability, useful in tasks like speech recognition or time-series analysis where interpretability is important but computational cost is low. \\
\hline
DeepSeek R1 7B (LLM) & 1 & 5 & 5 & Suitable for detailed long reasoning tasks where high Interpretability is required. However, its high computational cost makes it less ideal for real-time applications. Best for in-depth analysis and decision-making where reasoning transparency is crucial. \\
\hline
Ollama 3.1 8B (LLM) & 1 & 5 & 5 & Similar to DeepSeek, this LLM is suitable for tasks requiring long, detailed reasoning but is computationally expensive. Best for complex decision-making or language understanding tasks where transparency is important, but real-time performance is not a primary concern. \\
\hline
\end{tabular}

\begin{table}[H]
    \centering
    \caption{Numerical Rating Criteria for Accuracy, Interpretability, and Computational Cost}
    \label{tab:criteria}
    \resizebox{\textwidth}{!}{ 
    \begin{tabular}{|l|p{3cm}|p{5cm}|p{5cm}|}
        \hline
        \textbf{Metric} & \textbf{1 Star} & \textbf{2--3 Stars} & \textbf{4--5 Stars} \\
        \hline
        Accuracy & $<$ 0.6 & 0.6--0.8 & $>$ 0.8 \\
        \hline
        Interpretability & 
        The model is a "black box," and it is impossible to determine which specific input features (such as particular words, sentences, or attributes) led to the decision. These models typically only provide embedding vectors or class labels, but the reasoning behind the decisions is unclear. 
        & 
        The model provides some Interpretability but may not be as fully transparent as highly interpretable models. It is possible to identify which features, words, or sentences had an impact on the model's decision, but the full decision process may not be fully revealed. 
        & 
        The model is fully transparent, and the decision-making process is entirely understandable. Through Interpretability analysis, it is possible to clearly identify which features or words directly caused the classification or prediction result.
        \\
        \hline
        Computational Cost & Fast Inference $<$ 0.01s  & Moderate Inference 0.01 - 1s
 & Slow Inference $>$ 1s \\
        \hline
    \end{tabular}
    }
\end{table}

\section{Expert Validation}

To validate the experiment results identified in this study, we convened a virtual focus group composed of large language model agents in expert roles. We leverage LLM agent based focus group to produce diverse, high-quality viewpoints akin to human experts. We use this to discuss and cross-validates the trade-offs under investigation ~\cite{liu2024focusagent}\cite{zhang2024focus}.

\paragraph{Design and Prompting Protocol} In our experimental setup, we instantiated several AI agents, each assuming a distinct professional role relevant to AI-enabled medical device development and regulation. The assigned roles included a Regulatory Affairs Specialist (with FDA experience), a Quality Assurance Officer from a medical device company, an AI Engineer in the healthcare sector, a Clinical Expert specializing in AI diagnostic tools, a Compliance Officer familiar with EU MDR requirements, a Healthcare Data Scientist, a Regulatory Consultant experienced with NMPA and CE marking pathways, and a Medical Device Startup CEO focusing on AI-powered diagnostics. Each agent operated under role-specific system prompts to ensure that their responses reflected realistic domain knowledge and regulatory perspectives.


\paragraph{LLM Agent Prompting Strategy.}
To ensure that the responses generated by large language model (LLM) agents simulate expert reasoning rather than generic outputs, we implemented a structured prompting framework based on role-specific system instructions. Each agent was assigned a distinct regulatory or technical role—such as “Regulatory Affairs Specialist (FDA)” or “Compliance Officer (EU MDR)”—and received a dedicated system prompt instructing them to answer as if possessing real-world domain experience. This prompt design enforces persona grounding, encouraging responses that reflect the expectations, language, and decision logic of each stakeholder group.

Technically, we instantiated the discussion using the GPT-4 API with a consistent set of moderated questions across five rounds (warm-up, general trade-offs, model evaluation, explainability, and wrap-up). The system message provided to each agent followed the format:

\begin{quote}
\textsl{You are [Agent Role]. Please answer each question based on your real-world experience and role.}
\end{quote}

This was followed by a user prompt asking the substantive question under discussion. All interactions were logged and analyzed thematically. The complete focus group discussion logic, including role assignments, system message templates, and the discussion guide, is made available in the supplementary appendix and in our public repository\footnote{\url{https://github.com/RegAItool/classification-computational-complexity}}.

This prompting approach aligns with emerging best practices in multi-agent simulation using LLMs ~\cite{liu2024focusagent, zhang2024focus}, and allows us to elicit structured, role-consistent reasoning from LLMs that mimics stakeholder deliberation in regulatory affairs. The focus group session was moderated by a moderator, following a structured discussion guide divided into five sequential rounds. The session began with a warm-up introduction, allowing each agent to describe their role and primary concerns regarding AI-driven medical device classification. This was followed by a general discussion on initial perceptions of trade-offs among accuracy, Interpretability, and computational efficiency. Subsequent rounds focused on evaluating specific AI model types based on comparative performance data, examining the role of Interpretability in fostering regulatory trust, and reflecting on broader adoption barriers and opportunities. Each agent responded sequentially to the moderator's questions, and all dialogue was captured and stored as structured transcript data for subsequent analysis.

\section{Results}

\subsection{Model Tradeoff}

Insights from the focus group underscore the practical necessity of balancing model accuracy with Interpretability and operational constraints. 

\begin{figure}[H]
   \centering
    \includegraphics[width=1\textwidth]{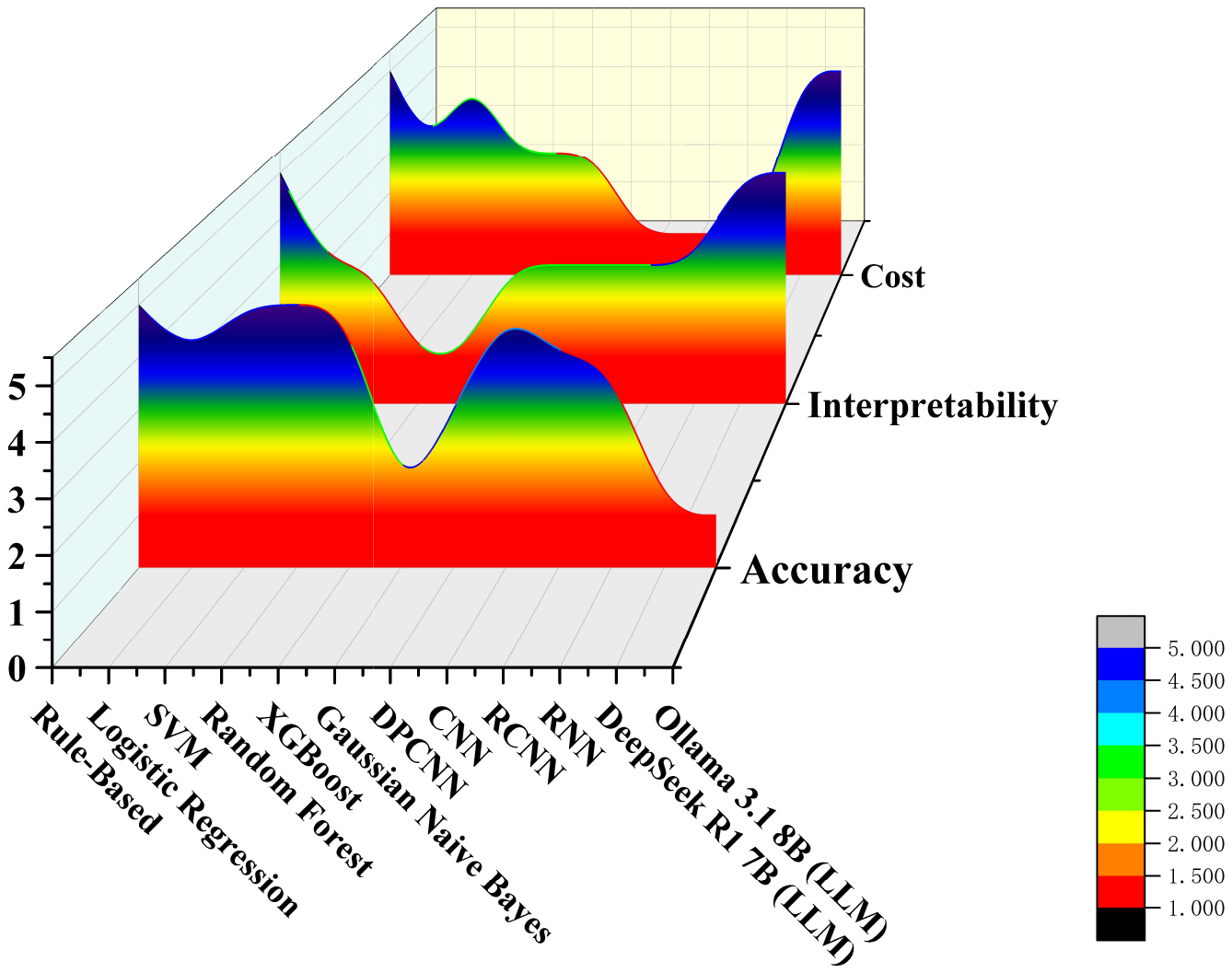}
    \caption{Comparison of Models Based on Accuracy, Interpretability, and Cost}
    \label{fig:llmmm}
\end{figure}

We showed each agent our experiment numerical results about each model accuracy, Interpretability and cost. To ensure comparability across models, we applied Min-Max normalization to all performance and efficiency metrics, scaling each variable to a range between 0 and 1. For instance, in the case of Interpretability, which was subjectively scored on a 1-to-5 scale, we normalized the values by subtracting the minimum score (1) and dividing by the range (4), resulting in a standardized value between 0 and 1. This approach allowed us to quantitatively compare models with varying performance characteristics, such as SVM, Logistic Regression, and Deep Learning models, while eliminating the bias introduced by differing scales of the original metrics. By using this method, we ensured that all metrics contributed equally to the subsequent analysis, facilitating a more accurate assessment of model efficiency and interpretability in this Figure \ref{fig:llmmm}. 

\subsection{Thematic Analysis}
\begin{figure}[H]
   \centering
    \includegraphics[width=0.9\textwidth]{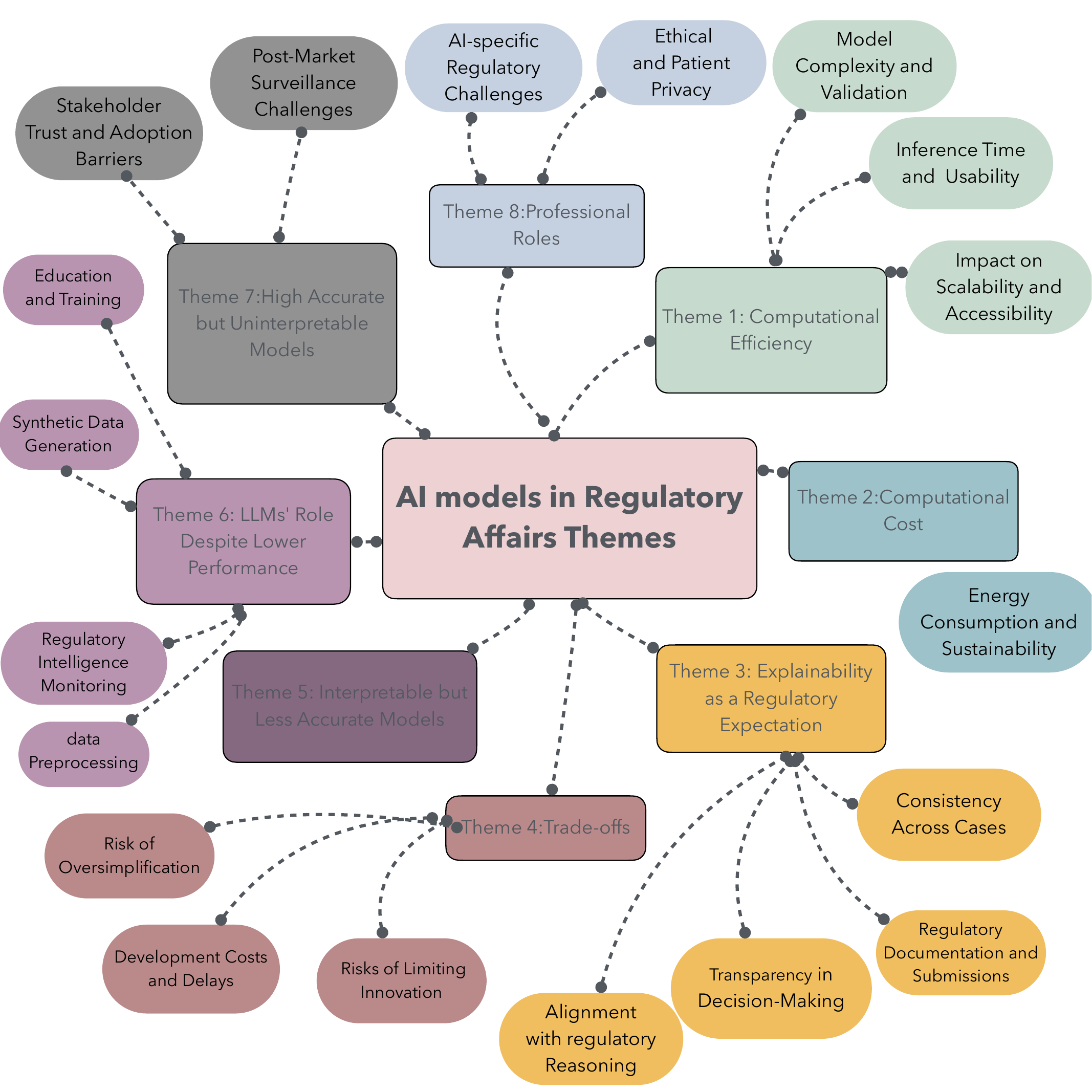}
    \caption{Theme Analysis of Models}
    \label{fig:theme_tree}
\end{figure}

To extract structured insights from our virtual focus group, we conducted a thematic analysis on the full LLM-generated transcript, following established qualitative coding protocols used in virtual expert simulations~\cite{liu2024focusagent}. Two independent researchers iteratively reviewed and annotated the responses to identify dominant themes and sub-codes. These themes were consolidated into eight overarching areas, presented in Figure~\ref{fig:theme_tree}, encompassing concerns such as computational efficiency, regulatory expectations for interpretability, and stakeholder trust.

Importantly, these themes were not abstract reflections on AI ethics or regulatory AI in general—they directly validated or challenged specific results from our empirical evaluations. For instance, the theme “High Accuracy but Limited Interpretability” emerged in response to the comparative performance of CNNs and XGBoost. Multiple agents expressed discomfort with deploying CNNs in high-risk contexts despite their superior accuracy (88.28\%), citing the lack of intuitive explanation mechanisms for convolutional filters. Similarly, XGBoost—while performant (accuracy 83.95\%, F1 score 0.64)—was criticized for its opaque ensemble structure, echoing our interpretability analysis that gradient-boosted trees, though powerful, lack transparency without tools like TreeSHAP, which are not always regulator-friendly or easily understood.

Conversely, under “Interpretable but Less Accurate Models,” agents endorsed the use of Logistic Regression in early-stage classification or legal audits, appreciating its transparent weight coefficients and compatibility with explanation tools like SHAP and Anchors. These insights supported our hybrid strategy proposal: using fast and interpretable models for initial triage or verification, even when not achieving the top predictive performance.

Themes around “LLMs’ Role Despite Lower Performance” also aligned with our experimental finding that models like DeepSeek (F1: 0.25) generated plausible reasoning but often failed to match legally correct classifications. Participants viewed LLMs as valuable for augmenting human insight—for example, generating prompts, structuring risk rationales, or supporting junior reviewers—rather than serving as standalone classifiers.

\section{Discussion and Conclusion}

This study systematically evaluated a range of machine learning and deep learning methods for the classification of medical devices, emphasizing the trade-offs between predictive accuracy and interpretability. Our findings offer insights not only into the computational performance of each model class, but also into their suitability for high-stakes regulatory contexts where Interpretability is essential.

\paragraph{Model Performance}
Support Vector Machines (SVMs) delivered reliable performance on structured embeddings and raw text, especially under well-separated class distributions. However, their lack of native interpretability and sensitivity to kernel selection render them less ideal for regulatory settings ~\cite{boser1992training}. Naive Bayes models, while computationally efficient and interpretable via log-likelihood inspection, assume conditional independence—a limitation when dealing with correlated language features typical in medical descriptions. Tree-based models exhibited distinct strengths. Decision Trees are transparent but prone to instability; in contrast, Random Forests and XGBoost significantly improve generalization through ensembling and boosting, respectively. When paired with TreeSHAP ~\cite{lundberg2020local}, these models offer both high accuracy and strong local/global interpretability, making them highly suitable for medical classification tasks that demand accountability.

Neural network models, including CNNs and RNNs, outperformed simpler models on complex inputs, capturing subtle semantic and syntactic patterns. Yet their black-box nature and sensitivity to input perturbations challenge post-hoc explanation tools like LIME and Integrated Gradients ~\cite{sundararajan2017axiomatic}. Our experiments confirmed that even token-level explanations can vary under slight changes, aligning with recent concerns about explanation stability ~\cite{alvarez2018robustness}. Transformer-based Large Language Models (LLMs) such as DeepSeek offer state-of-the-art performance in language understanding, but at the cost of interpretability and computational overhead. Despite LIME-Token approximations and attention map visualizations, their behavior remains opaque, echoing critiques that current methods may fall short of faithful explanation ~\cite{jacovi2020towards}.

While a 5\% difference in overall accuracy may seem negligible in aggregate metrics, its practical consequences can be substantial when disaggregated by class. For example, the Naïve Bayes model achieves a macro accuracy of only 59\%, yet its Class III F1 score reaches 0.63—far lower than CNN (0.81) and XGBoost (0.85). More strikingly, its F1 score for Class I is as low as 0.05, indicating an almost complete failure to identify low-risk devices correctly. In a large-scale deployment scenario processing 10,000 devices, this could lead to hundreds of Class I devices being incorrectly escalated into higher regulatory categories, burdening both manufacturers and regulators with unnecessary review processes. Conversely, its misclassification of Class III devices as lower risk poses serious safety and legal risks. This demonstrates that even modest differences in average performance may obscure critical weaknesses in specific classes, particularly when class distributions are imbalanced or when the cost of misclassification is not symmetric across regulatory categories. Therefore, per-class F1 scores offer a more meaningful lens than macro-averages alone for understanding real-world trade-offs in regulatory decision-making.

Beyond presenting aggregate metrics such as accuracy or macro F1, it is essential to contextualize these trade-offs in terms of real-world regulatory consequences. For example, a 5\% difference in overall accuracy between CNN (88.3\%) and XGBoost (84.0\%) translates to 50 additional misclassifications per 1,000 devices. However, the severity of this gap depends on which classes are most affected. As shown in our per-class evaluation, models with lower overall performance may still exhibit reasonable recall for high-risk Class III devices (e.g., Logistic Regression Class III F1 = 0.80), but fail entirely on low-risk Class I devices (F1 = 0.12). This imbalance can lead to misallocations of regulatory attention—either through under-regulating high-risk devices or overburdening low-risk ones. These risks are asymmetric and carry different legal and safety implications, further reinforcing the importance of per-class F1 scores as a core diagnostic metric.

\paragraph{Interpretability Nuances in Regulatory Contexts.}
Interpretability is not monolithic, and different models offer varying kinds of transparency. Logistic Regression provides interpretable decision boundaries via feature coefficients, allowing regulators to see which input terms directly influenced predictions—a valuable asset for traceability during audits. Models like XGBoost rely on post-hoc methods such as SHAP to estimate feature importance. While these techniques are powerful, they can be challenging for non-technical reviewers to interpret, as the explanations often reference abstract embedding dimensions rather than specific, identifiable words that directly trigger a given classification. LLMs, on the other hand, generate natural language rationales that may appear intuitive and persuasive, but do not always align with legally valid reasoning. This presents a nuanced trade-off: explanations must not only be understandable but also legally and procedurally correct. A natural-language explanation that is plausible but fails to cite a critical regulatory criterion may increase trust while introducing hidden error. Therefore, the trade-off is not merely accuracy versus interpretability, but accuracy versus relevant and valid interpretability.

\paragraph{Computational Cost Beyond Inference Time.}
Although our experimental comparison emphasizes inference time as a core item, other components of computational cost deserve consideration. Training and fine-tuning costs—particularly for LLMs and deep neural networks—can be prohibitive, requiring specialized hardware and extensive optimization. Moreover, preparing high-quality annotated datasets for regulatory domains is labor-intensive and requires domain expertise, increasing the up-front cost of model development. Maintenance is also non-trivial; models must be regularly validated against evolving regulatory standards and product taxonomies. Finally, the infrastructure needed to deploy high-complexity models (e.g., LLMs) may require cloud-based GPUs, which pose additional data privacy concerns and increase ongoing expenses. In contrast, simpler models like Logistic Regression or XGBoost can often be deployed on secure, local hardware, making them more viable for agencies with limited resources or strict data governance requirements. As such, computational cost must be viewed holistically, encompassing not just speed but also sustainability, infrastructure compatibility, and maintainability over time.

\paragraph{Regulatory Considerations/Proposed Framework for Evaluating Explanation Utility in Regulatory Affairs}
Regulatory agencies such as the FDA, EMA, and NMPA increasingly emphasize Interpretability in AI-assisted healthcare. The European Union’s AI Act mandates transparency for high-risk systems, and the FDA’s Good Machine Learning Practice (GMLP) framework similarly requires explainable outputs ~\cite{aiact2021,fda_gmlp}. Our use of SHAP, Anchors, and LIME supports these expectations, offering as a benchmark study for model deployment in regulatory contexts.

While the lack of a universally accepted benchmark for interpretability is well-documented in AI research, regulatory contexts demand more than abstract notions of explanation fidelity. In the domain of medical device regulation, interpretability is not just an academic ideal—it is a practical necessity that underpins legal compliance, accountability, and trust in automated decision-making. To address this, we propose a domain-specific framework for evaluating the utility of explanations in regulatory affairs, structured around three interrelated dimensions. \textbf{Comprehensibility} refers to whether the explanation is presented in a form and language that regulators can easily understand. For example, highlighting tokens such as ``implantable'' or ``single-use'' may have immediate regulatory significance, while references to latent embedding dimensions or heatmaps without clear verbal context may offer little actionable value. \textbf{Justifiability} asks whether the explanation provides a traceable rationale aligned with legal or procedural standards. Ideally, explanations should reflect established decision heuristics in risk classification—such as identifying claims of AI functionality, intended use in critical care, or the presence of control boxes that imply greater system complexity. \textbf{Trust Alignment} concerns whether the explanation increases a regulator’s confidence in the model’s decision-making. This can be assessed by comparing model outputs with expert judgments or regulatory documentation. For instance, explanations that consistently reference device attributes cited in official classification guidelines (e.g., YY/T 1753-2020) are more likely to foster trust and support regulatory adoption.

We recommend that future work in regulatory AI consider formalizing these dimensions into scoring rubrics, decision protocols, or hybrid review workflows. For example, combining a high-accuracy model like CNN or XGBoost with a justifiable explanation layer (e.g., TreeSHAP or Anchors) may offer a balance of performance and regulatory traceability. This would help ensure that AI tools do not merely offer statistical predictions but generate legally and procedurally coherent justifications—a prerequisite for use in regulated environments.

\paragraph{Reframing Trade-offs Through Misclassification Risk and Scenario-Based Stress Testing.}

While model evaluation typically revolves around the trade-offs between accuracy, interpretability, and computational cost, these technical objectives must ultimately be reframed in terms of regulatory outcomes. In high-stakes domains such as medical device oversight, the primary cost of algorithmic failure is not computational inefficiency or model opacity, but rather the misclassification of devices in ways that compromise patient safety or hinder innovation.

We propose incorporating the \textsl{cost of misclassification} into the narrative of model evaluation. A model that marginally reduces inference time or memory usage may still be unsuitable for deployment if it increases the risk of under-regulating a high-risk (Class III) device or wrongly rejecting a safe and compliant product. For instance, our evaluation shows that although Naïve Bayes is computationally efficient, it performs poorly on Class I (F1: 0.05) and significantly underperforms on Class III (F1: 0.63) compared to CNN (F1: 0.81) and XGBoost (F1: 0.85). Such deficiencies could result in hundreds of devices per 10,000 being incorrectly categorized, leading to unnecessary regulatory burden or, worse, compromised safety assurance. Therefore, accuracy differences must be interpreted through the lens of per-class risk exposure.

To make this trade-off analysis more tangible, we advocate for \textsl{scenario-based stress testing}. This involves constructing hypothetical but realistic classification scenarios and examining how different model types perform under these conditions. For example:

\begin{quote}
\textbf{Scenario:} A manufacturer submits a device description for a novel AI-powered diagnostic tool employing an untested algorithm. The textual description is linguistically complex and includes subtle, ambiguous claims about its intended use and performance. Based on the information provided, the device could plausibly be assigned to Class II or Class III.

\textbf{Implication:} In this case, the cost of misclassification is asymmetrical: incorrectly assigning the device to Class II could bypass critical review stages, while a Class III designation might delay market entry unjustifiably. Traditional models may lack the semantic flexibility to interpret nuanced claims, while deep learning or LLM-based systems may provide richer language understanding. However, these models differ in their ability to offer transparent, audit-ready justifications for their predictions. In this scenario, interpretability becomes paramount—not only to trace model decisions, but also to provide regulators with human-comprehensible rationales for why the device is high or medium risk.
\end{quote}

This approach moves beyond static evaluation metrics and grounds the model comparison in the actual decision-making pressures faced by regulatory authorities. It highlights that model acceptability is contingent not only on accuracy or latency but also on alignment with domain-specific risk tolerance and interpretability needs. We propose future regulatory AI studies adopt such stress testing methodologies to better capture real-world implications and stakeholder priorities.

\paragraph{On the Gap Between Language Plausibility and Regulatory Correctness.}
While large language models (LLMs) offer fluent and human-like justifications for classification decisions, our findings underscore a crucial limitation for regulatory applications: their outputs may exhibit language plausibility without satisfying legal correctness. For instance, as shown in Figure~\ref{fig:llmm}, an LLM classified a Class III device as Class II, despite articulating a seemingly well-reasoned explanation. This divergence arises because LLMs are trained on general language data and do not inherently encode jurisdiction-specific definitions—such as the NMPA’s explicit requirement that Class III devices typically involve life-sustaining, life-supporting, or high-risk implantable functions.

This distinction matters in regulatory settings, where classification errors can have significant legal and safety implications. Unlike domain-trained models or symbolic systems explicitly grounded in rule-based logic, LLMs lack guaranteed alignment with formal regulatory taxonomies. As such, we propose that LLMs be deployed only in constrained or supervised roles, such as generating human-interpretable rationales or assisting in post hoc review—never as autonomous decision-makers. Embedding structured regulatory rules into prompts, or integrating LLMs with rule-based or expert-reviewed pipelines, may help mitigate these limitations while still leveraging their generative strength.

\paragraph{Decision Tree Based on Scenarios}
No single model excels across all criteria, underscoring the need for context-dependent strategies. For example, a hybrid approach could employ a high-accuracy model for initial screening, followed by a simpler, more interpretable model or human expert review for high-risk cases. 

\begin{figure}[H]
   \centering
\includegraphics[width=0.65\textwidth]{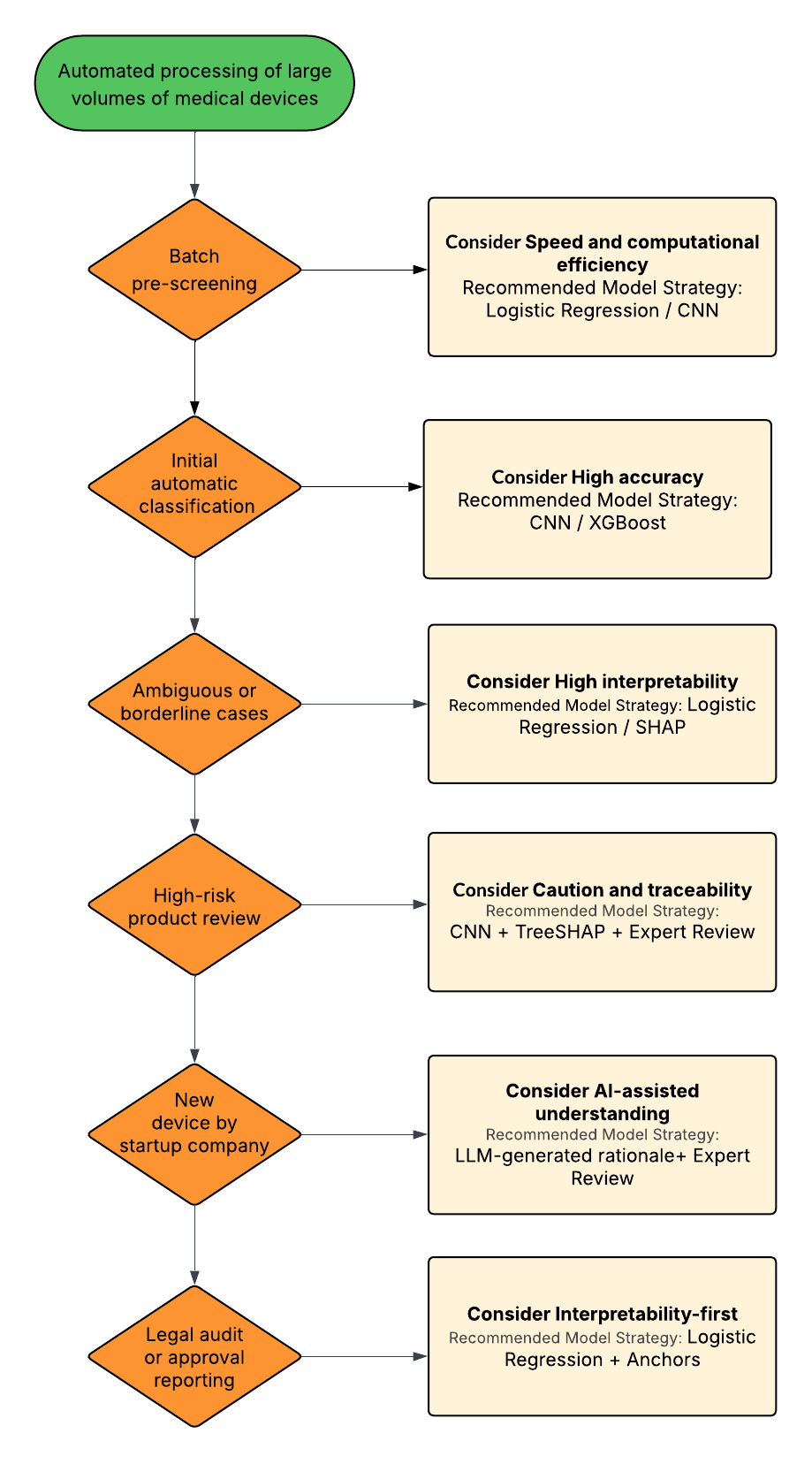}
    \caption{AI Algorithm Decision Tree in Different Regulatory Affairs Scenarios}
    \label{fig:tree}
\end{figure}

This is real-world regulatory scenario workflows as shown in Figure \ref{fig:tree}, AI model selection is often contingent on the stage of evaluation and the specific demands of each decision point. This begins with batch pre-screening, where the objective is to efficiently process a large volume of devices by authority using lightweight and fast models such as logistic regression or CNNs. These models balance moderate accuracy with low computational overhead, making them well-suited for rapid triage.

Following pre-screening, initial automatic classification is conducted using higher-accuracy models such as CNN or XGBoost. These models are capable of capturing complex linguistic patterns in product descriptions and are ideal for generating a preliminary risk class prediction at scale.

For ambiguous cases, where model confidence is low or multiple risk classes are nearly equally probable, the workflow transitions to interpretable models such as logistic regression combined with SHAP explanations. This ensures that regulatory reviewers can understand the basis of the model’s decision and intervene when needed. In the case of high-risk product reviews—such as potential Class III devices—additional caution and traceability are required. Here, we recommend using CNNs combined with TreeSHAP explanations and expert human review, enabling both predictive accuracy and accountability in regulatory justification. When evaluating a novel device submitted by a startup company, regulatory experts may benefit from LLMs to generate natural language rationales, particularly in the absence of precedent or structured features. LLMs can synthesize device descriptions into explainable summaries that facilitate regulatory interpretation. During legal audits or final approval documentation, where interpretability is paramount, simple yet transparent models such as logistic regression combined with Anchors explanations are most appropriate. These allow reviewers and legal stakeholders to trace exactly how classification decisions were made, fulfilling traceability and documentation requirements.

\paragraph{Limitations and Future Directions.}
This work is constrained by three key factors: (1) Modelling with real-world registries, (2) challenges in objectively evaluating interpretability quality, and (3) regulatory ambiguity surrounding AI model approval pathways. While we benchmarked LIME, SHAP, Anchors, and Integrated Gradients, their comparative trustworthiness remains difficult to quantify. Future efforts should consider hybrid models that combine symbolic reasoning and deep learning, or apply causal Interpretability frameworks for better robustness ~\cite{wachter2017counterfactual}. (4) In the future, researchers could leverage feature engineering techniques—such as the regulatory text features identified in this study—to improve model generalizability, enhance performance on underrepresented device categories, and increase interpretability. For example, incorporating features based on regulatory keywords, device risk indicators, or functional descriptors could help models better align with domain-specific reasoning and support more transparent decision-making in regulatory workflows. This study did not incorporate such features, as our goal was to benchmark model performance without the influence of additional feature engineering.

For future work, domain-specific interpretability scoring frameworks could be developed in the future, incorporating regulatory heuristics such as alignment with official classification rules (e.g., presence of risk-indicative terms like "implantable", "disposable", or "AI-driven"). These frameworks could assess whether model explanations adhere to known regulatory decision-making patterns. There is potential of expert-in-the-loop evaluation protocols, where human regulatory professionals validate explanations generated by AI models. This could involve structured annotation tasks, comparative assessments, or consensus scoring, thereby grounding interpretability evaluation in real-world regulatory reasoning. Regulatory-aligned explanation benchmarks, similar to how benchmark datasets such as MIMIC-III have enabled systematic development and evaluation of clinical NLP models~\cite{johnson2016mimic}. Such benchmarks would contain annotated examples of device descriptions, regulatory decisions, and justifications, enabling standardized evaluation of model outputs for faithfulness and utility.

\section{Conflicts of Interest}
The authors affirm that there are no conflicts of interest to declare. 

\section*{DATA AVAILABILITY}
The scripts used to carry out the analysis and the resultant data that support the findings in this study are available on: 
https://github.com/RegAItool/classification-computational-complexity


\bibliographystyle{IEEEtran}
\bibliography{od_prediction}   

\begin{thebibliography}{10}
\providecommand{\url}[1]{#1}
\csname url@samestyle\endcsname
\providecommand{\newblock}{\relax}
\providecommand{\bibinfo}[2]{#2}
\providecommand{\BIBentrySTDinterwordspacing}{\spaceskip=0pt\relax}
\providecommand{\BIBentryALTinterwordstretchfactor}{4}
\providecommand{\BIBentryALTinterwordspacing}{\spaceskip=\fontdimen2\font plus
\BIBentryALTinterwordstretchfactor\fontdimen3\font minus \fontdimen4\font\relax}
\providecommand{\BIBforeignlanguage}[2]{{%
\expandafter\ifx\csname l@#1\endcsname\relax
\typeout{** WARNING: IEEEtran.bst: No hyphenation pattern has been}%
\typeout{** loaded for the language `#1'. Using the pattern for}%
\typeout{** the default language instead.}%
\else
\language=\csname l@#1\endcsname
\fi
#2}}
\providecommand{\BIBdecl}{\relax}
\BIBdecl

\bibitem{FDA2025}
\BIBentryALTinterwordspacing
{U.S. Food and Drug Administration}, ``Fda announces completion of first ai-assisted scientific review pilot and aggressive agency-wide ai rollout,'' May 2025, accessed: 2025-05-10. [Online]. Available: \url{https://www.fda.gov/news-events/press-announcements/fda-announces-completion-first-ai-assisted-scientific-review-pilot-and-aggressive-agency-wide-ai}
\BIBentrySTDinterwordspacing

\bibitem{Heaven2025}
\BIBentryALTinterwordspacing
W.~D. Heaven, ``Openai and the fda are talking about ai for drug evaluation,'' \emph{WIRED}, May 2025, accessed: 2025-05-10. [Online]. Available: \url{https://www.wired.com/story/openai-fda-doge-ai-drug-evaluation/}
\BIBentrySTDinterwordspacing

\bibitem{mantus2014fda}
D.~Mantus and D.~J. Pisano, \emph{FDA regulatory affairs}.\hskip 1em plus 0.5em minus 0.4em\relax CRC Press, 2014.

\bibitem{kumari2016current}
B.~S. Kumari, G.~S. Hanuja, M.~Nagabhushanam, D.~N. Reddy, and B.~Bonthagarala, ``Current regulatory requirements for registration of medicines, compilation and submission of dossier in australian therapeutic goods administration,'' \emph{International Journal of Advanced Scientific and Technical Research, ISSN}, pp. 2249--9954, 2016.

\bibitem{han2024regulatory}
Y.~Han, A.~Ceross, and J.~Bergmann, ``Regulatory frameworks for ai-enabled medical device software in china: Comparative analysis and review of implications for global manufacturer,'' \emph{JMIR AI}, vol.~3, p. e46871, 2024.

\bibitem{han2023uncovering}
Y.~Han, A.~Ceross, and J.~H. Bergmann, ``Uncovering regulatory affairs complexity in medical products: A qualitative assessment utilizing open coding and natural language processing (nlp),'' \emph{arXiv preprint arXiv:2401.02975}, 2023.

\bibitem{muehlematter2023fda}
U.~J. Muehlematter, C.~Bluethgen, and K.~N. Vokinger, ``Fda-cleared artificial intelligence and machine learning-based medical devices and their 510 (k) predicate networks,'' \emph{The Lancet Digital Health}, vol.~5, no.~9, pp. e618--e626, 2023.

\bibitem{aronson2020medical}
J.~K. Aronson, C.~Heneghan, and R.~E. Ferner, ``Medical devices: definition, classification, and regulatory implications,'' \emph{Drug safety}, vol.~43, no.~2, pp. 83--93, 2020.

\bibitem{pomeranz2013comprehensive}
J.~L. Pomeranz, ``A comprehensive strategy to overhaul fda authority for misleading food labels,'' \emph{American journal of law \& medicine}, vol.~39, no.~4, pp. 617--647, 2013.

\bibitem{herman2021position}
A.~Herman, W.~Uter, T.~Rustemeyer, M.~Matura, K.~Aalto-Korte, J.~Duus~Johansen, M.~Gon{\c{c}}alo, I.~R. White, A.~Balato, A.~M. Gim{\'e}nez~Arnau \emph{et~al.}, ``Position statement: the need for eu legislation to require disclosure and labelling of the composition of medical devices,'' \emph{Journal of the European Academy of Dermatology and Venereology}, vol.~35, no.~7, pp. 1444--1448, 2021.

\bibitem{arnould2021complexity}
A.~Arnould, R.~Hendricusdottir, and J.~Bergmann, ``The complexity of medical device regulations has increased, as assessed through data-driven techniques,'' \emph{Prosthesis}, vol.~3, no.~4, pp. 314--330, 2021.

\bibitem{han2024more}
Y.~Han, A.~Ceross, and J.~Bergmann, ``More than red tape: exploring complexity in medical device regulatory affairs,'' \emph{Frontiers in Medicine}, vol.~11, p. 1415319, 2024.

\bibitem{yin2021role}
J.~Yin, K.~Y. Ngiam, and H.~H. Teo, ``Role of artificial intelligence applications in real-life clinical practice: systematic review,'' \emph{Journal of medical Internet research}, vol.~23, no.~4, p. e25759, 2021.

\bibitem{castiglioni2021ai}
I.~Castiglioni, L.~Rundo, M.~Codari, G.~Di~Leo, C.~Salvatore, M.~Interlenghi, F.~Gallivanone, A.~Cozzi, N.~C. D'Amico, and F.~Sardanelli, ``Ai applications to medical images: From machine learning to deep learning,'' \emph{Physica medica}, vol.~83, pp. 9--24, 2021.

\bibitem{imaichi2013comparison}
O.~Imaichi, T.~Yanase, and Y.~Niwa, ``A comparison of rule-based and machine learning methods for medical information extraction,'' in \emph{The first workshop on natural language processing for medical and healthcare fields}, 2013, pp. 38--42.

\bibitem{patil2023artificial}
R.~S. Patil, S.~B. Kulkarni, and V.~L. Gaikwad, ``Artificial intelligence in pharmaceutical regulatory affairs,'' \emph{Drug Discovery Today}, p. 103700, 2023.

\bibitem{khinvasara2024post}
T.~Khinvasara, N.~Tzenios, and A.~Shanker, ``Post-market surveillance of medical devices using ai,'' \emph{Journal of Complementary and Alternative Medical Research}, vol.~25, no.~7, pp. 108--122, 2024.

\bibitem{kumar2010automatic}
D.~M. Kumar, ``Automatic induction of rule based text categorization,'' \emph{International Journal of Computer Science \& Information Technology (IJCSIT)}, vol.~2, no.~6, 2010.

\bibitem{khan2010review}
A.~Khan, B.~Baharudin, L.~H. Lee, and K.~Khan, ``A review of machine learning algorithms for text-documents classification,'' \emph{Journal of advances in information technology}, vol.~1, no.~1, pp. 4--20, 2010.

\bibitem{korger2021rule}
A.~Korger and J.~Baumeister, ``Rule-based semantic relation extraction in regulatory documents.'' in \emph{LWDA}, 2021, pp. 26--37.

\bibitem{awad2015support}
M.~Awad, R.~Khanna, M.~Awad, and R.~Khanna, ``Support vector machines for classification,'' \emph{Efficient learning machines: Theories, concepts, and applications for engineers and system designers}, pp. 39--66, 2015.

\bibitem{webb2010naive}
G.~I. Webb, E.~Keogh, and R.~Miikkulainen, ``Na{\"\i}ve bayes.'' \emph{Encyclopedia of machine learning}, vol.~15, no.~1, pp. 713--714, 2010.

\bibitem{salama2016semantic}
D.~M. Salama and N.~M. El-Gohary, ``Semantic text classification for supporting automated compliance checking in construction,'' \emph{Journal of Computing in Civil Engineering}, vol.~30, no.~1, p. 04014106, 2016.

\bibitem{zhang2016semantic}
J.~Zhang and N.~M. El-Gohary, ``Semantic nlp-based information extraction from construction regulatory documents for automated compliance checking,'' \emph{Journal of Computing in Civil Engineering}, vol.~30, no.~2, p. 04015014, 2016.

\bibitem{ceross2021machine}
A.~Ceross and J.~Bergmann, ``A machine learning approach for medical device classification,'' in \emph{Proceedings of the 14th International Conference on Theory and Practice of Electronic Governance}, 2021, pp. 285--291.

\bibitem{banerjee2023large}
S.~Banerjee, P.~Dunn, S.~Conard, and R.~Ng, ``Large language modeling and classical ai methods for the future of healthcare,'' \emph{Journal of Medicine, Surgery, and Public Health}, vol.~1, p. 100026, 2023.

\bibitem{wu2024regulating}
\BIBentryALTinterwordspacing
K.~Wu, E.~Wu, K.~Rodolfa, D.~E. Ho, and J.~Zou, ``Regulating ai adaptation: An analysis of ai medical device updates,'' \emph{arXiv preprint arXiv:2407.16900}, 2024. [Online]. Available: \url{https://arxiv.org/abs/2407.16900}
\BIBentrySTDinterwordspacing

\bibitem{shen2023towards}
\BIBentryALTinterwordspacing
X.~Shen, H.~Brown, J.~Tao, M.~Strobel, Y.~Tong, A.~Narayan, H.~Soh, and F.~Doshi-Velez, ``Towards regulatable ai systems: Technical gaps and policy recommendations,'' \emph{arXiv preprint arXiv:2306.12609}, 2023. [Online]. Available: \url{https://arxiv.org/abs/2306.12609}
\BIBentrySTDinterwordspacing

\bibitem{wang2025comprehensive}
\BIBentryALTinterwordspacing
e.~a. Wang, ``A comprehensive survey on on-device ai models,'' \emph{arXiv preprint arXiv:2503.06027}, 2025. [Online]. Available: \url{https://arxiv.org/abs/2503.06027}
\BIBentrySTDinterwordspacing

\bibitem{amari1999improving}
S.-i. Amari and S.~Wu, ``Improving support vector machine classifiers by modifying kernel functions,'' \emph{Neural networks}, vol.~12, no.~6, pp. 783--789, 1999.

\bibitem{chen2016xgboost}
T.~Chen and C.~Guestrin, ``Xgboost: A scalable tree boosting system,'' \emph{Proceedings of the 22nd ACM SIGKDD International Conference on Knowledge Discovery and Data Mining}, pp. 785--794, 2016.

\bibitem{lundberg2017unified}
S.~M. Lundberg and S.-I. Lee, ``A unified approach to interpreting model predictions,'' \emph{Advances in Neural Information Processing Systems}, vol.~30, 2017.

\bibitem{kim2014convolutional}
Y.~Kim, ``Convolutional neural networks for sentence classification,'' in \emph{Proceedings of the 2014 Conference on Empirical Methods in Natural Language Processing (EMNLP)}.\hskip 1em plus 0.5em minus 0.4em\relax Association for Computational Linguistics, 2014, pp. 1746--1751.

\bibitem{yin2017comparative}
W.~Yin, K.~Kann, M.~Yu, and H.~Sch{\"u}tze, ``Comparative study of cnn and rnn for natural language processing,'' \emph{arXiv preprint arXiv:1702.01923}, 2017.

\bibitem{johnson2017deep}
R.~Johnson and T.~Zhang, ``Deep pyramid convolutional neural networks for text categorization,'' in \emph{Proceedings of the 55th Annual Meeting of the Association for Computational Linguistics (Volume 1: Long Papers)}.\hskip 1em plus 0.5em minus 0.4em\relax Association for Computational Linguistics, 2017, pp. 562--570.

\bibitem{lai2015recurrent}
S.~Lai, L.~Xu, K.~Liu, and J.~Zhao, ``Recurrent convolutional neural networks for text classification,'' in \emph{Proceedings of the 29th AAAI Conference on Artificial Intelligence}, 2015, pp. 2267--2273.

\bibitem{sundararajan2017axiomatic}
M.~Sundararajan, A.~Taly, and Q.~Yan, ``Axiomatic attribution for deep networks,'' \emph{International Conference on Machine Learning (ICML)}, pp. 3319--3328, 2017.

\bibitem{ribeiro2016lime}
M.~T. Ribeiro, S.~Singh, and C.~Guestrin, ``Why should i trust you? explaining the predictions of any classifier,'' in \emph{Proceedings of the 22nd ACM SIGKDD International Conference on Knowledge Discovery and Data Mining}.\hskip 1em plus 0.5em minus 0.4em\relax ACM, 2016, pp. 1135--1144.

\bibitem{vaswani2017attention}
A.~Vaswani, N.~Shazeer, N.~Parmar, J.~Uszkoreit, L.~Jones, A.~N. Gomez, {\L}.~Kaiser, and I.~Polosukhin, ``Attention is all you need,'' in \emph{Advances in Neural Information Processing Systems}, vol.~30, 2017.

\bibitem{devlin2018bert}
J.~Devlin, M.-W. Chang, K.~Lee, and K.~Toutanova, ``Bert: Pre-training of deep bidirectional transformers for language understanding,'' \emph{arXiv preprint arXiv:1810.04805}, 2018.

\bibitem{cui2019pretraining}
Y.~Cui, W.~Che, T.~Liu, B.~Qin, S.~Wang, and G.~Hu, ``Pre-training with whole word masking for chinese bert,'' \emph{arXiv preprint arXiv:1906.08101}, 2019.

\bibitem{touvron2023llama}
H.~Touvron, T.~Lavril, G.~Izacard, X.~Martinet, M.-A. Lachaux, T.~Lacroix, B.~Rozi{\`e}re, N.~Goyal, E.~Hambro, F.~Azhar \emph{et~al.}, ``Llama: Open and efficient foundation language models,'' \emph{arXiv preprint arXiv:2302.13971}, 2023.

\bibitem{deepseek2024model}
D.~AI, ``Deepseek-vl: Scaling vision-language alignment with decoupled pretraining,'' \emph{arXiv preprint arXiv:2401.14185}, 2024.

\bibitem{jacovi2020towards}
A.~Jacovi and Y.~Goldberg, ``Towards faithfully interpretable nlp systems: How should we define and evaluate faithfulness?'' \emph{Proceedings of the 58th Annual Meeting of the Association for Computational Linguistics (ACL)}, pp. 4198--4205, 2020.

\bibitem{rudin2019stop}
C.~Rudin, ``Stop explaining black box machine learning models for high stakes decisions and use interpretable models instead,'' \emph{Nature Machine Intelligence}, vol.~1, no.~5, pp. 206--215, 2019.

\bibitem{ribeiro2018anchors}
M.~T. Ribeiro, S.~Singh, and C.~Guestrin, ``Anchors: High-precision model-agnostic explanations,'' \emph{Proceedings of the AAAI Conference on Artificial Intelligence}, vol.~32, no.~1, 2018.

\bibitem{lundberg2018explainable}
S.~M. Lundberg, B.~Nair, M.~Vavilala, M.~Horibe, M.~J. Eisses, T.~Adams, D.~E. Liston, D.~K. Low, S.-F. Newman, J.~Kim \emph{et~al.}, ``Explainable machine-learning predictions for the prevention of hypoxaemia during surgery,'' \emph{Nature Biomedical Engineering}, vol.~2, no.~10, pp. 749--760, 2018.

\bibitem{lundberg2020local}
S.~M. Lundberg, G.~Erion, H.~Chen, A.~DeGrave, J.~M. Prutkin, B.~Nair, R.~Katz, J.~Himmelfarb, N.~Bansal, and S.-I. Lee, ``Local explanation methods for tree-based models: A unified approach,'' \emph{Nature Machine Intelligence}, vol.~2, no.~1, pp. 56--67, 2020.

\bibitem{strubell2019energy}
E.~Strubell, A.~Ganesh, and A.~McCallum, ``Energy and policy considerations for deep learning in nlp,'' \emph{arXiv preprint arXiv:1906.02243}, 2019.

\bibitem{wu2022survey}
Z.~Wu, X.~Liu, D.~F. Zhou, Y.~He, and J.~Gao, ``A survey on efficient training of transformers,'' \emph{arXiv preprint arXiv:2202.06930}, 2022.

\bibitem{sanh2019distilbert}
V.~Sanh, L.~Debut, J.~Chaumond, and T.~Wolf, ``Distilbert, a distilled version of bert: smaller, faster, cheaper and lighter,'' \emph{arXiv preprint arXiv:1910.01108}, 2019.

\bibitem{zhang2024towards}
C.~Zhang, Y.~Liu, X.~Yang, K.~Gu, J.~Shen, X.~Li, and J.~Bian, ``Towards practical trade-offs between interpretability and performance: A composite interpretability metric for trustworthy ai,'' 2024.

\bibitem{kaur2022trust}
H.~Kaur, M.~Desai, M.~Eslami, and J.~Forlizzi, ``Trust and transparency in human-ai interaction: A survey of trust calibration, user understanding, and interpretability,'' in \emph{Proceedings of the 2022 ACM Conference on Fairness, Accountability, and Transparency (FAccT)}, 2022, pp. 487--498.

\bibitem{chen2024taxonomy}
S.~Chen, T.~Wu, and Q.~V. Lee, ``A taxonomy of interpretability in human-centered ai: From explanations to user experience,'' \emph{IEEE Transactions on Visualization and Computer Graphics}, 2024.

\bibitem{nmpa_udi_database}
{National Medical Products Administration}, ``Udi database,'' \url{https://udi.nmpa.gov.cn/download.html}, n.d., accessed: 2024-09-25.

\bibitem{tasos2019stratified}
\BIBentryALTinterwordspacing
Tasos, ``Split the data between the training data and test data using sklearn,'' 2019, accessed: 2025-05-04. [Online]. Available: \url{https://datascience.stackexchange.com/questions/51236/split-the-data-between-the-training-data-and-test-data-using-sklearn}
\BIBentrySTDinterwordspacing

\bibitem{scikit-learn}
F.~Pedregosa, G.~Varoquaux, A.~Gramfort, V.~Michel, B.~Thirion, O.~Grisel, M.~Blondel, P.~Prettenhofer, R.~Weiss, V.~Dubourg, J.~Vanderplas, A.~Passos, D.~Cournapeau, M.~Brucher, M.~Perrot, and {\'E}.~Duchesnay, ``Scikit-learn: Machine learning in python,'' \emph{Journal of Machine Learning Research}, vol.~12, pp. 2825--2830, 2011.

\bibitem{liu2024focusagent}
A.~F. N. L.~N. Liu, ``Focusagent: [article title],'' \emph{Journal Name}, vol. Volume Number, no. Issue Number, p. Page Range, 2024.

\bibitem{zhang2024focus}
T.~Zhang, X.~Zhang, R.~Cools, and A.~Simeone, ``Focus agent: Llm-powered virtual focus group,'' in \emph{Proceedings of the 24th ACM International Conference on Intelligent Virtual Agents}, 2024, pp. 1--10.

\bibitem{boser1992training}
B.~E. Boser, I.~M. Guyon, and V.~N. Vapnik, ``A training algorithm for optimal margin classifiers,'' \emph{Proceedings of the fifth annual workshop on Computational learning theory}, pp. 144--152, 1992.

\bibitem{alvarez2018robustness}
D.~Alvarez-Melis and T.~S. Jaakkola, ``On the robustness of interpretability methods,'' \emph{arXiv preprint arXiv:1806.08049}, 2018.

\bibitem{aiact2021}
E.~Commission, ``Proposal for a regulation on a european approach for artificial intelligence,'' 2021, available at: \url{https://eur-lex.europa.eu/legal-content/EN/TXT/?uri=CELEX:52021PC0206}.

\bibitem{fda_gmlp}
U.~Food and D.~A. (FDA), ``Good machine learning practice for medical device development: Guiding principles,'' 2021, available at: \url{https://www.fda.gov/medical-devices/software-medical-device-samd/good-machine-learning-practice-medical-device-development-guiding-principles}.

\bibitem{wachter2017counterfactual}
S.~Wachter, B.~Mittelstadt, and L.~Floridi, ``Counterfactual explanations without opening the black box: Automated decisions and the gdpr,'' \emph{Harvard journal of law \& technology}, vol.~31, p. 841, 2017.

\bibitem{johnson2016mimic}
A.~E. Johnson, T.~J. Pollard, L.~Shen, L.-w.~H. Lehman, M.~Feng, M.~Ghassemi, B.~Moody, P.~Szolovits, L.~A. Celi, and R.~G. Mark, ``Mimic-iii, a freely accessible critical care database,'' \emph{Scientific data}, vol.~3, p. 160035, 2016.

\end{thebibliography}
\end{document}